\documentclass[lettersize,journal]{IEEEtran}
\usepackage{amsmath,amsfonts}
\usepackage{array}
\usepackage[caption=false,font=normalsize,labelfont=sf,textfont=sf]{subfig}
\usepackage{textcomp}
\usepackage{stfloats}
\usepackage{url}
\usepackage{verbatim}
\usepackage{graphicx}
\usepackage{soul}
\usepackage[table]{xcolor}
\usepackage{balance}
\usepackage{cite}
\usepackage{algorithm}
\usepackage{booktabs}
\usepackage{multirow}
\usepackage{listings}
\usepackage{tikz,siunitx}
\usepackage{times}
\usepackage{helvet}
\usepackage{tcolorbox}
\tcbuselibrary{listings}
\usepackage{courier}
\usepackage{afterpage}
\usepackage{adjustbox}
\usepackage{algpseudocode}
\usepackage{caption}
\usepackage{hyperref}

\definecolor{myblue}{HTML}{407cb9}     
\definecolor{myred}{HTML}{e26062}      
\definecolor{mygreen}{HTML}{61a658}    
\definecolor{myorange}{HTML}{F57C00}  
\definecolor{mypurple}{HTML}{7B1FA2}   

\def\BibTeX{{\rm B\kern-.05em{\sc i\kern-.025em b}\kern-.08em
    T\kern-.1667em\lower.7ex\hbox{E}\kern-.125emX}}

\begin{document}
\title{MeEvo: Metacognitive Evolution Combined with Natural Evolution for Automatic Heuristic Design}
\author{Zishang Qiu\textsuperscript{\rm 1}, 
    Xinan Chen\textsuperscript{\rm 1},
    Rong Qu\textsuperscript{\rm 2},
    Ruibin Bai\textsuperscript{\rm 1}\\
    \textsuperscript{\rm 1}School of Computer Science, University of Nottingham Ningbo China, Ningbo, China \\
    \textsuperscript{\rm 2}School of Computer Science, University of Nottingham, Nottingham, UK\\
    \texttt{xinan.chen@nottingham.edu.cn}
}
\maketitle

\begin{abstract}
Large Language Models (LLMs) have advanced Automatic Heuristic Design (AHD) by enabling heuristic generation through reasoning and code synthesis. In LLM-based AHD, the LLM reasons about algorithm design and generates executable heuristic code. Existing architectures adopt two main paradigms: Natural Evolution applies crossover and mutation to this code to explore diverse strategies, but discards the reasoning traces behind the design decisions, weakening knowledge retention; Metacognitive Evolution retains these reasoning traces and refines them through reflection, but lacks population-level recombination, limiting exploration. These limitations reduce search efficiency, stability, and solution quality on complex problems. To address this gap, we propose MeEvo, an AHD framework that cyclically couples Natural Evolution and Metacognitive Evolution with operator balance that shifts from exploration to exploitation. Natural Evolution explores heuristic code while recording LLM-generated reasoning traces, fitness values, errors and best heuristic into a shared history; Metacognitive Evolution then reflects on this history to generate improved heuristics that feed into the next Natural Evolution cycle. This design enables population-driven exploration and reflection-driven refinement to reinforce each other. Experiments on five optimization problems show that MeEvo achieves stronger performance and lower variance than tested LLM-based AHD architectures, especially on complex constrained tasks.
\end{abstract}


\section{Introduction}
\label{sec:intro}
\IEEEPARstart{M}{any} real-world problems, from logistics and resource allocation to network design, require optimization heuristics: rules of thumb that find good solutions efficiently \cite{kim1993heuristic}. Designing such heuristics demands extensive expertise and is inherently problem-specific; the No Free Lunch theorem \cite{wolpert1997no} further guarantees that no single heuristic works best for all problems, requiring repeated manual effort for each new task. Automatic Heuristic Design (AHD) \cite{sabar2014automatic} aims to automate this process, with classical approaches ranging from Genetic Programming \cite{espejo2009survey} to reinforcement learning \cite{yi2022automated}; however, these remain constrained by human-defined search spaces \cite{liu2024evolution}.

Large Language Models (LLMs) \cite{zhao2026survey} have transformed AHD by combining evolutionary computation with neural code generation \cite{zhang2024understanding}. LLM-generated heuristics inherently comprise two coupled components: \textit{reasoning traces}, the LLM's explanations of its design choices that record design knowledge, and executable code subject to fitness evaluation. Existing LLM-based AHD architectures pursue two main paradigms, each with distinct strengths. \textbf{Natural Evolution} architectures (FunSearch \cite{romera2024mathematical}, EoH \cite{liu2024evolution}, ReEvo \cite{ye2024reevo}, MCTS-AHD \cite{zheng2025monte}) apply population-driven crossover and mutation to explore the executable-code space broadly, generating diverse algorithmic strategies through recombination and variation. However, while some capture reasoning traces as byproducts (EoH) or employ auxiliary reflection to guide search (ReEvo), none preserve reasoning as independently reusable information across generations. The strategic knowledge behind effective code is lost between cycles. \textbf{Metacognitive Evolution} (MeLA \cite{QIU2026133022}) adopts a fundamentally different strategy. Rather than evolving heuristic code directly, MeLA evolves the LLM-generated reasoning traces that record the design rationale behind each heuristic. After evaluation, metacognitive reflection analyzes these traces against their performance outcomes, identifying which reasoning patterns correlate with effective heuristics and which lead to failure, and distills these insights to generate improved heuristics that feed into the next generation. Through this iterative reflection, the LLM progressively refines its design logic across generations, but without population-driven crossover and mutation it cannot discover fundamentally different algorithmic strategies outside its accumulated reasoning trajectory. These two paradigms exhibit \textit{complementary deficiencies}: Natural Evolution explores diverse executable code but discards the design knowledge in reasoning traces, while Metacognitive Evolution accumulates such traces but lacks the code diversity that crossover and mutation provide. Since these two components must be jointly exploited, neither paradigm alone realizes the full potential of LLM-generated heuristics.

Separately, prompt-level architectures (OPRO \cite{yang2024large}, EvoPrompt \cite{guo2023evoprompt}, HiFo-Prompt \cite{chen2025hifo}, EvoPH \cite{liu2025experience}) evolve the task descriptions fed to the LLM rather than its outputs, and therefore operate at a different level from the two core paradigms above. Recent hybrid architectures such as EoH-MR \cite{qi2025memetic} and PyVRP\textsuperscript{+} \cite{malik2026pyvrp} have begun combining EC with metacognitive reflection, but in both cases reflection remains an auxiliary module rather than an independent evolutionary process; neither couples population-driven crossover and mutation with reflection-driven metacognition as equal, alternating partners.

This paper introduces \textbf{MeEvo}, a framework that combines Natural Evolution and Metacognitive Evolution through three contributions:

\textbf{(1) Cyclic alternation of Natural Evolution and Metacognitive Evolution.}
Natural Evolution and Metacognitive Evolution alternate cyclically and communicate through shared history (detailed in \S\ref{subsec:overview}), rather than co-evolving in parallel as in prior architectures.

\textbf{(2) Hierarchical operator balance without external coordination.}
We achieve exploration-exploitation balance at two levels without external coordination: the alternation rhythm between the two processes governs macro-level balance, while within Natural Evolution, the crossover-to-mutation ratio adapts autonomously from exploration-dominated to exploitation-focused as the search progresses.

\textbf{(3) Comprehensive experimental validation across diverse optimization problems.}
We validate MeEvo on five problems spanning classical benchmarks, black-box optimization, and real-world constrained tasks, with ablation studies, parameter sensitivity analysis, and statistical significance testing demonstrating the framework's effectiveness and generalizability.

\begin{figure}[htbp]
    \centering
    \includegraphics[width=\columnwidth]{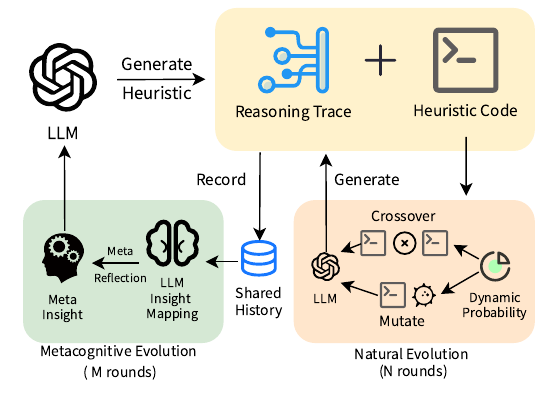}
    \caption{Architecture of MeEvo combining Natural Evolution and Metacognitive Evolution. Reasoning traces record design choices, while heuristic code is evaluated for fitness. The two processes alternate and cooperate through shared history.}
    \label{compare}
\end{figure}

\textbf{Fig.~\ref{compare}} illustrates this architecture.

Code and Results link: https://github.com/Qzs1335/MeEvo.

\section{Literature Review}
\label{sec:lit_review}

\subsection{Natural Evolution for AHD: Crossover, Mutation, and Population-Driven Search}
\label{subsec:natural_evo_lit}

Combining LLMs with evolutionary computation has produced a new generation of AHD architectures \cite{liu2024evolution}. A central thread applies \textit{Natural Evolution} operators (crossover, mutation, and selection acting on a population) to explore the heuristic code space. FunSearch \cite{romera2024mathematical} constrained LLM-generated code within a predefined skeleton and used selection to retain top performers. EoH \cite{liu2024evolution} introduced LLM-mediated crossover and mutation, co-evolving natural-language thoughts alongside heuristic code. ReEvo \cite{ye2024reevo} enhanced these operators with dual-level reflection that analyzes parent pairs to guide crossover and accumulates expertise across iterations to guide mutation. MCTS-AHD \cite{zheng2025monte} replaced the population with a Monte Carlo tree, preserving underperforming candidates as revisitable nodes for more thorough exploration. EoH-S \cite{liu2026eoh} extended EoH by evolving a set of heuristics rather than a single best candidate, maintaining population diversity through heuristic-set-level selection. Zhang et al.~\cite{zhang2025llm} instead addressed instance heterogeneity by partitioning problem classes into feature-based sub-classes and generating tailored heuristics for each---a direction orthogonal to population-level evolution. A parallel line of work evolves the prompts guiding code generation (OPRO \cite{yang2024large}, EvoPrompt \cite{guo2023evoprompt}, HiFo-Prompt \cite{chen2025hifo}, EvoPH \cite{liu2025experience}) rather than the code itself; these architectures address a different challenge and are summarized in \textbf{TABLE~\ref{tab:comparison}}.

Despite their success, the Natural Evolution architectures share a structural limitation: they apply Natural Evolution operators exclusively to heuristic code. Reasoning traces produced alongside code generation are treated as ephemeral byproducts. Even when explicitly captured, as in EoH and ReEvo, reasoning is not preserved as independently reusable information across generations. Crossover and mutation inherit code fragments but not the strategic knowledge behind why a particular component proved effective. Consequently, population-driven exploration discovers diverse strategies but cannot accumulate design knowledge over evolutionary cycles.

\subsection{Metacognitive Evolution for AHD: Reflection-Driven Refinement}
\label{subsec:metacog_evo_lit}

A complementary thread targets the LLM's reasoning process itself through \textit{metacognitive reflection}: the capacity to monitor, evaluate, and regulate one's own cognitive processes \cite{lai2011metacognition}. This is distinct from reasoning enhancement techniques such as Chain-of-Thought (CoT) \cite{NEURIPS2022_9d560961} and Tree-of-Thought (ToT) \cite{yao2023tree}, which structure \textit{how} the model thinks but do not govern \textit{whether} its current strategy is effective. Metacognition is a higher-order regulatory layer: the ability to step back and assess whether the current reasoning trajectory is productive \cite{lin2025think,wang2024metacognitive,li2025adaptive}.

MeLA \cite{QIU2026133022} instantiated metacognition for AHD through Metacognitive Evolution. Rather than evolving heuristic code through crossover and mutation, or task descriptions through forward-looking optimization, MeLA evolves the LLM's reasoning traces by metacognitively reflecting on previously generated heuristic design, execution errors, and fitness outcomes. Its metacognitive search engine analyzes these records to reinforce effective reasoning patterns, preserve high-performing algorithmic components from the best heuristic, and hypothesize novel improvement directions, then distills these insights to guide subsequent heuristic generation. However, MeLA operates as a single-process system: heuristic generation and metacognitive reflection are coupled within one loop, without the diversity that population-driven crossover and mutation provide. While metacognitive reflection progressively deepens the LLM's design logic, it cannot independently discover fundamentally novel algorithmic strategies outside its accumulated reasoning trajectory.

Emerging hybrid architectures represent early steps toward combining these two paradigms. EoH-MR \cite{qi2025memetic} augments EoH with CMA-ES local optimization and a meta-cognitive reflection module that monitors population diversity, fitness improvement velocity, and stagnation status, then uses these signals to dynamically adjust operator selection probabilities (e.g., promoting exploratory mutation when diversity collapses) and prune low-yield prompt templates. PyVRP\textsuperscript{+} \cite{malik2026pyvrp} introduces a Reason-Act-Reflect cycle for metacognitive heuristic evolution. However, metacognitive reflection remains an auxiliary module serving code generation rather than an independent evolutionary process. In EoH-MR, meta-cognitive monitoring operates on the population as a guidance signal; in PyVRP\textsuperscript{+}, the Reason-Act-Reflect cycle refines individual solutions but does not alternate between population-driven exploration and metacognitive reflection as independent evolutionary processes. Consequently, neither provides a principled mechanism that treats reasoning as a separately maintained entity with dedicated refinement operators, bidirectionally coupled with code-level evolution.

The progression from Natural Evolution architectures to Metacognitive Evolution, with recent hybrid efforts beginning to bridge them (\textbf{TABLE~\ref{tab:comparison}}), reveals a consistent pattern. Even in the most advanced architectures, population-driven crossover and mutation and reflection-driven metacognition have never been combined as equal, alternating partners in a unified framework.

\begin{table*}[htbp]
\centering
\renewcommand{\arraystretch}{1.25}
\caption{Evolution Paradigm Comparison in LLM-based AHD}
\label{tab:comparison}
\resizebox{\textwidth}{!}{%
\begin{tabular}{@{}cccc@{}}
\toprule
\textbf{Method} & \textbf{Evolution Paradigm} & \textbf{Key Mechanism} & \textbf{Reasoning as Independent Process} \\
\midrule
FunSearch \cite{romera2024mathematical} & Natural Evolution & Selection + skeleton constraint & No \\
EoH \cite{liu2024evolution} & Natural Evolution & Crossover + mutation (5 strategies) & No \\
ReEvo \cite{ye2024reevo} & Natural Evolution & Dual-level reflection-enhanced crossover/mutation & No \\
MCTS-AHD \cite{zheng2025monte} & Natural Evolution & Tree search (population alternative) & No \\
EoH-S \cite{liu2026eoh} & Natural Evolution & Heuristic-set-level selection & No \\
\midrule
Zhang et al.~\cite{zhang2025llm} & Instance-specific AHD & Subclass partitioning + tailored heuristic generation & No \\
\midrule
EvoPrompt \cite{guo2023evoprompt} & Prompt-level & EC operators on task descriptions & No \\
OPRO \cite{yang2024large} & Prompt-level & Training-set evaluation & No \\
HiFo-Prompt \cite{chen2025hifo} & Prompt-level & Navigator rule engine & No \\
EvoPH \cite{liu2025experience} & Prompt-level & Policy sampling + Island migration & No \\
\midrule
MeLA \cite{QIU2026133022} & Metacognitive Evolution & Metacognitive reflection & Yes (single-process) \\
\midrule
EoH-MR \cite{qi2025memetic} & NE + auxiliary reflection & Meta-cognitive monitoring + CMA-ES & No \\
PyVRP\textsuperscript{+} \cite{malik2026pyvrp} & NE + auxiliary reflection & Reason-Act-Reflect cycle & No \\
\midrule
\textbf{MeEvo (Ours)} & \textbf{NE + ME (cyclic alternation)} & \textbf{Process alternation + operator balance} & \textbf{Yes} \\
\bottomrule
\end{tabular}%
}
\end{table*}

\subsection{Exploration-Exploitation Mechanisms in AHD}
\label{subsec:ee_review}

A central challenge in optimization is balancing \textit{exploration} (searching broadly for new promising regions) against \textit{exploitation} (refining solutions in known good regions) \cite{kennedy1995particle, hussain2019exploration}. In classical algorithms, this balance is controlled by numerical parameters, such as crossover and mutation rates in GA \cite{holland1992genetic}, inertia weights in PSO \cite{kennedy1995particle}, or temperature in SA, but the optimal setting varies by problem and changes during the search.

LLM-based AHD introduces a new possibility: using the model's reasoning capabilities for higher-level, semantic exploration-exploitation decisions rather than relying solely on low-level numerical parameters. Existing architectures adopt distinct strategies (\textbf{TABLE~\ref{tab:ee_detail}}): HiFo-Prompt uses a Navigator rule engine; EvoPH uses policy sampling and island migration; EoH-MR uses meta-cognitive monitoring. All rely on external coordination mechanisms.

MeEvo differs from parallel co-evolution architectures such as EvoPH \cite{liu2025experience}: rather than running two populations simultaneously with explicit coupling mechanisms (policy sampling, island migration), MeEvo cyclically alternates between Natural Evolution and Metacognitive Evolution. The rationale and implications of this design are detailed in \S\ref{subsec:overview}.

\begin{table*}[htbp]
\centering
\renewcommand{\arraystretch}{1.3}
\caption{Comparison of Cyclic Alternation Architectures in LLM-based AHD}
\label{tab:ee_detail}
\resizebox{\textwidth}{!}{%
\begin{tabular}{@{}ccccc@{}}
\toprule
\textbf{Method} & \textbf{Process 1} & \textbf{Process 2} & \textbf{Relationship} & \textbf{Exploration--Exploitation Control} \\
\midrule
HiFo-Prompt \cite{chen2025hifo} & Prompt design & Code generation & Single-process + dual-module & Navigator rule engine \\
EvoPH \cite{liu2025experience} & Prompt template & Code generation & Parallel co-evolution & Policy sampling + Island migration \\
\midrule
\textbf{MeEvo (Ours)} & \textbf{Natural Evolution (crossover+mutation)} & \textbf{Metacognitive Evolution (reflection)} & \textbf{Cyclic alternation} & \textbf{Hierarchical (process alternation + operator balance)} \\
\bottomrule
\end{tabular}%
}
\end{table*}

\subsection{Summary and Identified Research Gap}
\label{subsec:research_gap}

The literature reveals three interrelated gaps. \textbf{Gap 1: Natural Evolution and Metacognitive Evolution have never been systematically combined.} EC-based architectures (FunSearch, EoH, ReEvo, MCTS-AHD) apply population-driven crossover and mutation to explore code space, but discard reasoning traces and therefore cannot accumulate design knowledge. MeLA applies metacognitive reflection to reason about design choices, but lacks population-driven diversity. Recent hybrid architectures (EoH-MR, PyVRP\textsuperscript{+}) incorporate reflection as an auxiliary module serving code generation; treating NE and ME as equal, alternating partners has not been investigated. \textbf{Gap 2: No architecture separates reasoning and code as distinct entities with dedicated processing mechanisms.} Reasoning is not preserved as a separately maintained entity with its own refinement operators, despite the benefits of separating design rationale from executable code. \textbf{Gap 3: No architecture achieves exploration-exploitation balance through structural process alternation.} Existing mechanisms (Navigator rule engines in HiFo-Prompt, policy sampling in EvoPH, meta-cognitive monitoring in EoH-MR) rely on external coordination; a balance governed by the alternation rhythm itself has not been explored.

MeEvo addresses these three gaps through the design presented next.
 
\section{Methodology}

\subsection{System Overview}
\label{subsec:overview}

MeEvo combines two complementary evolutionary paradigms, \textbf{Natural Evolution} and \textbf{Metacognitive Evolution}, through cyclic alternation. The optimization proceeds through alternation of two processes:

\begin{itemize}
    \item \textbf{Natural Evolution:} The LLM applies population-driven crossover and mutation directly to heuristic code. An operator balance shifts from crossover-dominated exploration early in search to mutation-focused exploitation later. Every generated code variant is accompanied by a \textit{reasoning trace}, the LLM's explanation of its design choices, which is recorded for later metacognitive analysis.

    \item \textbf{Metacognitive Evolution:} The LLM reflects on the accumulated \textit{shared history}: reasoning traces, fitness values, execution errors, and the best heuristic, recorded across past generations to produce \textit{Meta Insights}: structured reflections that diagnose convergence, identify ineffective strategies, and propose algorithmic improvement pathways. These insights are then translated into new heuristic code, which joins the population to seed the next Natural Evolution cycle.
\end{itemize}

\begin{figure*}[t]
    \centering
    \includegraphics[width=\textwidth]{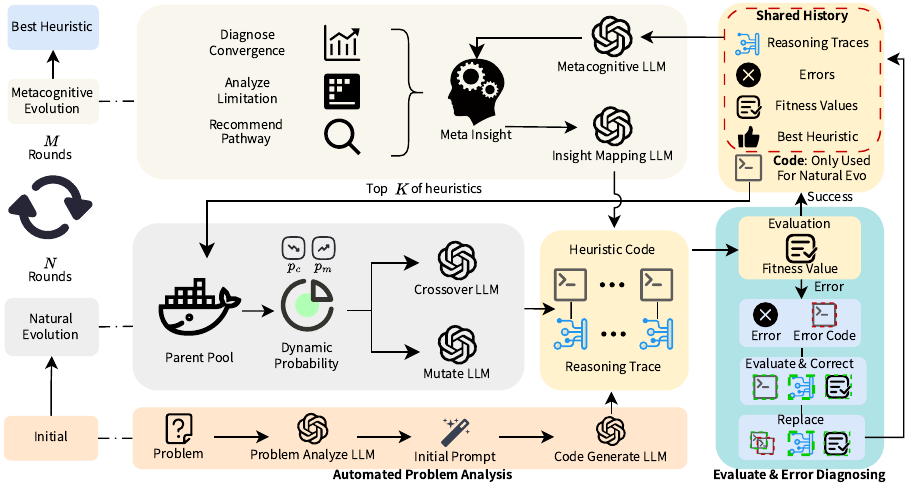}
    \caption{Overall procedure of MeEvo. Natural Evolution and Metacognitive Evolution alternate cyclically, coupled through shared history.}
    \label{fig:flowchart}
\end{figure*}

A natural question is why MeEvo alternates between Natural Evolution and Metacognitive Evolution rather than running them in parallel, as EvoPH \cite{liu2025experience} does for prompts and code. The two processes have incompatible operational requirements. Natural Evolution applies crossover to recombine parent heuristics, which requires a stable parent pool; if Metacognitive Evolution injected new heuristics mid-cycle, the reference set would shift and crossover would become unreliable. Metacognitive Evolution analyzes the shared history accumulated across Natural Evolution generations and requires traces spanning many generations to distinguish genuine patterns from noise; a few recent traces cannot provide this. Cyclic alternation satisfies both requirements: Natural Evolution first explores the code space with a fixed parent pool over $N$ generations, then Metacognitive Evolution reflects on the accumulated history over $M$ generations and produces improved heuristics that re-enter the parent pool for the next cycle. This design differs from parallel co-evolution in three respects: the two processes alternate rather than run simultaneously, giving each a stable operating environment; their relationship is asymmetric (Natural Evolution produces reasoning traces as byproducts of code generation, Metacognitive Evolution consumes them to produce refinements); and exploration-exploitation balance is governed by the alternation rhythm rather than by external coordination mechanisms.

\subsection{Iteration Steps}
\label{subsec:iteration}

The architecture takes as input the target problem $P$, initial population size $N_0$, per-generation size $N_{pop}$, evaluation budget $L$, correction limit $COR$, parent pool size $K$, and cycle counts $N$ and $M$. Each heuristic is evaluated by executing it on the target problem instances to compute the objective value (lower is better). The main loop alternates $N$ Natural Evolution generations with $M$ Metacognitive Evolution generations until the evaluation budget $L$ is exhausted, returning the global best $h^\ast$. A consolidated summary of all symbols is provided in the Appendix.

For example, with $N{=}2$ and $M{=}1$ (the default configuration validated in \S\ref{subsec:ablation}): Natural Evolution runs two generations of crossover and mutation, each producing $N_{pop}$ offspring and recording their reasoning traces into shared history. Metacognitive Evolution then reflects on this accumulated history to generate improved heuristics, which are added to the population $H$. The next Natural Evolution cycle selects parents from this enriched pool, so metacognitively refined heuristics participate in crossover alongside existing ones. This NE$\rightarrow$ME$\rightarrow$NE rhythm repeats until the evaluation budget $L$ is exhausted.

The \textit{shared history} comprises four record types accumulated across all generations: reasoning traces $TH$, fitness values $F$, execution errors $ERR$, and the best heuristic $h^\ast$.

\textbf{Algorithm~\ref{alg:main}} formalizes the main loop. The framework initializes the evaluation counter, the population $H$, three historical record sets ($F$, $TH$, $ERR$), and the global best $h^\ast$. The problem is first analyzed (\textsc{Analyze}) to produce a structured description $APA$, which guides generation of $N_0$ initial heuristics. The main loop then alternates between \textbf{Natural Evolution} and \textbf{Metacognitive Evolution}. Natural Evolution selects the top-$K$ feasible heuristics as parents, applies LLM-mediated crossover and mutation with operator balance (Eq.~\ref{eq:pcross}), and records each offspring's reasoning trace. After each Metacognitive Evolution phase, the improved heuristics it produces enter the population, giving the next Natural Evolution cycle higher-quality starting material for crossover and mutation. Metacognitive Evolution reflects on accumulated shared history to generate Meta Insights, which an insight-mapping step translates into $N_{pop}$ new heuristics. Each heuristic is evaluated and corrected for execution errors as detailed in the Appendix. We assign $f(h)=\infty$ only for execution failures; otherwise a finite value is returned. The loop continues until budget $L$ is exhausted, returning the global best $h^\ast$.

\begin{algorithm}[htbp]
\caption{Main Framework of Cyclic Alternation}
\label{alg:main}
\small
\begin{algorithmic}[1]
\setlength{\itemsep}{0pt}
\State \textbf{Input:} $P$, $N_0$, $N_{pop}$, $L$, $COR$, $N$, $M$, $K$
\State \textbf{Output:} Best heuristic $h^\ast$, fitness $f(h^\ast)$
\State $Eval \!\gets\! 0$, $H\!\gets\!\emptyset$, $F\!\gets\!\emptyset$, $TH \!\gets\! \emptyset$, $ERR \!\gets\! \emptyset$, $f(h^\ast) \!\gets\! \infty$
\State $APA \gets \textsc{Analyze}(P)$
\State Generate $N_0$ initial heuristics via \textsc{InitHeuristic}; evaluate and update $H, F, TH, ERR, h^\ast$

\While{$Eval < L$} \Comment{Budget-limited cyclic loop}
    \State \textbf{// --- Natural Evolution ---}
    \For{$i \gets 1$ \textbf{to} $N$}
        \State $\mathcal{P} \gets$ top-$K$ feasible heuristics from $H$ sorted by $f(h)$
        \State Compute $p_c$, $p_m$ via Eq.~\ref{eq:pcross}
        \State Generate $N_{pop}$ offspring via \textsc{LLMCrossover}/\textsc{LLMMutate}
        \State Evaluate offspring, update $H, F, TH, ERR, h^\ast$, $Eval$
    \EndFor
    \State \textbf{// --- Metacognitive Evolution ---}
    \State $Meta \gets \textsc{MetacognitiveReflection}(TH, F, ERR, h^\ast)$
    \For{$j \gets 1$ \textbf{to} $M$}
        \State Generate $N_{pop}$ heuristics from $Meta$; evaluate and update $H, F, TH, ERR, h^\ast$, $Eval$
    \EndFor
\EndWhile

\State \Return $h^\ast, f(h^\ast)$
\end{algorithmic}
\end{algorithm}

The auxiliary algorithms and typical execution errors encountered during heuristic evaluation are detailed in the Appendix.

\subsection{Natural Evolution: Population-Driven Crossover and Mutation}
\label{subsec:nature_evo}

In the Natural Evolution process, the LLM acts as an evolutionary operator, applying population-driven crossover and mutation directly to heuristic code (see Appendix for the full procedure). Crossover recombines heterogeneous parent strategies; mutation introduces local code modifications. Together, these two operators expand the frontier of algorithmic strategies explored. Reasoning traces produced alongside each generation are recorded in shared history for subsequent metacognitive reflection. The entire population is replaced by offspring each generation. Each iteration executes three steps:

\subsubsection{Parent Selection}
Infeasible heuristics ($f(h) = \infty$) are excluded. The remaining feasible heuristics $\mathcal{H}_{v}$ are sorted by ascending fitness, and the top $K$ form the parent pool $\mathcal{P}$. If fewer than two feasible parents exist, the iteration is skipped.

\subsubsection{LLM-Mediated Crossover}
With probability $p_c$, two distinct parents $h_a, h_b \in \mathcal{P}$ are selected, and \textsc{LLMCrossover} prompts the LLM to: (1) analyze each parent's distinct algorithmic approach, (2) identify complementary strengths, (3) synthesize a hybrid offspring that combines components in novel ways, and (4) introduce at least one enhancement not present in either parent. By recombining heterogeneous strategies, crossover explores regions of the heuristic space inaccessible to single-parent variation. The LLM outputs a reasoning trace $Th_{off}$ that records which parent contributed each component and why those components were combined.

\subsubsection{LLM-Mediated Mutation}
With probability $p_m = 1 - p_c$, a single parent is sampled, and \textsc{LLMMutate} prompts the LLM to modify the heuristic through local code changes. Illustrative examples are provided for guidance: adaptive step-size tuning, Cauchy perturbation, and chaotic local search for continuous optimization; 2-opt/3-opt, swap, and inversion operators for combinatorial optimization. The LLM is not restricted to these named techniques and is instructed to focus on refining search mechanisms rather than merely tuning numerical parameters; this semantic-level guidance incurs additional token cost relative to random operators (see \S\ref{subsec:cost}).

\subsubsection{Operator Balance}
The crossover and mutation probabilities shift linearly with search progress:
\begin{equation}
p_c(\text{Eval}) = \alpha + \beta \times \left(1 - \frac{\text{Eval}}{L}\right)
\label{eq:pcross}
\end{equation}
\begin{equation}
p_m(\text{Eval}) = 1 - p_c(\text{Eval})
\label{eq:pmut}
\end{equation}
where $Eval$ is the current number of function evaluations, $L$ is the maximum budget, $\alpha$ is the baseline crossover probability when the budget nears exhaustion ($Eval/L{\to}1$), and $\beta$ is the adaptation amplitude, with $\alpha{+}\beta \leq 1$. The constraint $p_c + p_m = 1$ ensures each offspring is generated by exactly one operator, giving the linear shift direct control over the exploration--exploitation mix in each generation. Our default setting uses $\alpha{=}0.5$ and $\beta{=}0.2$ (see Section~\ref{sec:experiments} for sensitivity analysis), yielding $p_c$ ranging from $0.7$ at the start of search to $0.5$ near budget exhaustion. Crossover dominates early, emphasizing \textbf{exploration} by recombining diverse parent strategies, while mutation probability increases as the search progresses, shifting the balance toward \textbf{exploitation}. Empirically, this setting outperforms the alternatives (\S\ref{subsec:param_sensitivity}).

\subsection{Metacognitive Evolution: Reflection-Driven Refinement}
\label{subsec:reasoning_evo}

Whereas Natural Evolution explores the code space through stochastic variation, Metacognitive Evolution refines the search through \textit{directed reflection} on the shared history. The LLM first analyzes the history of reasoning traces and their fitness outcomes, then translates these strategic reflections into improved heuristics. This process embodies the principle of metacognition: stepping back from individual code-generation steps to evaluate and regulate the overall search strategy.

\subsubsection{Two-Step Reflection-to-Code Pipeline}
The process occurs in two steps. First, the metacognitive LLM distills the shared history (reasoning history $TH$, fitness history $F$, error records $ERR$, and current best $h^\ast$) into a Meta Insight:
\begin{equation}
\text{Meta} = \mathcal{R}(TH, F, ERR, h^\ast)
\label{eq:meta}
\end{equation}
where $\mathcal{R}$ is the metacognitive reflection function. The Meta Insight is a structured natural-language report containing three components: \textit{Convergence Diagnosis} (classifying the search state as premature convergence, steady improvement, oscillation, or stagnation), \textit{Limitation Analysis} (identifying cognitive limitations and repeated ineffective strategies), and \textit{Pathway Recommendation} (proposing specific optimization pathways with named algorithmic techniques).

Second, the insight mapping LLM translates the Meta Insight into executable code with a reasoning trace:
\begin{equation}
h_{t+1} = \mathcal{G}(\text{Meta}, APA), \quad Th_{t+1} = \mathcal{T}(\text{Meta})
\label{eq:mapping}
\end{equation}
where $\mathcal{G}$ generates code and $\mathcal{T}$ generates the reasoning trace. Separating reflection from code generation into two steps is important: the Meta Insight captures \textit{what} should change and \textit{why} at a strategic level, while insight mapping determines \textit{how} to implement these changes in code. This separation ensures that the reflection is grounded in accumulated history rather than being influenced by the immediate code generation task, and it allows the Meta Insight to serve as an interpretable, reusable signal across multiple code generation iterations. The insight mapping step prompts the LLM to preserve the best heuristic's strongest components while redesigning weak areas identified in the Meta Insight, producing code grounded in accumulated strategic knowledge rather than generated from scratch. The reasoning history remains an immutable record, while metacognitive reflection generates forward-looking improvement signals from it.

\subsubsection{Convergence-Aware Reflection and Evolutionary Operators}
The Meta Insight performs three tasks. \textit{Convergence Diagnosis} classifies the search state (premature convergence, steady improvement, oscillation, or stagnation) by analyzing fitness trends in $F$ and error patterns in $ERR$. \textit{Limitation Analysis} identifies cognitive biases and repeated ineffective strategies. \textit{Pathway Recommendation} proposes specific improvement pathways with named algorithmic techniques. These tasks provide directed variation that complements the stochastic variation of crossover and mutation in Natural Evolution. Shared history ($TH$, $F$, $ERR$, $h^\ast$) carries reasoning traces and fitness outcomes forward across cycles.

\subsection{Shared Components and Prompt Design}
\label{subsec:enhanced}

Three components share design principles with the metacognitive reflection approach described in MeLA \cite{QIU2026133022} and are enhanced for the cyclic alternation architecture: automated problem analysis, error diagnosis, and metacognitive reflection (extended from single-process reflection to Meta Insights with convergence diagnosis, limitation analysis, and pathway recommendation). All LLM calls use structured prompts with defined roles and output formats; complete prompts are provided in the Appendix. The overall procedure is illustrated in \textbf{Fig.~\ref{fig:flowchart}}.

\section{Experiments}
\label{sec:experiments}

\subsection{Challenges in Real-World Problems}

A critical question is whether LLM-based AHD generalizes beyond classical benchmarks. We include two real-world problems, ACS and WSN, to test this: they demand reasoning about domain-specific constraints beyond generic algorithmic templates, testing whether AHD architectures remain effective when LLMs lack domain-specific knowledge. Existing architectures perform poorly here without structured problem analysis; incomplete problem descriptions yield near-zero runnable heuristics for architectures such as EoH and ReEvo (see Appendix), making evolutionary search infeasible.

\subsection{Experimental Setup}

We evaluate MeEvo on five optimization problems, categorized by evaluation protocol: two downstream-solver benchmarks (TSP-ACO, BPP-ACO), two real-world constrained problems (ACS, WSN), and one constructive-heuristic benchmark (TSP-Construct), using two LLM backbones, DeepSeek-V4-Flash and MIMO-v2.5-Pro, selected for their strong open-source performance, API availability, and low inference cost. Related settings are provided in \textbf{TABLE~\ref{tab:llm_config}}. The problems are:

\begin{itemize}

    \item \textbf{TSP with ACO framework (TSP-ACO):} The same TSP problem, but the LLM generates an $(n \times n)$ heuristic matrix that guides an Ant Colony Optimization process \cite{ye2024reevo}. The matrix is clipped to $[10^{-9}, 10^6]$ and normalized before use by ACO (30 ants, 100 iterations, $\rho{=}0.9$, $\alpha_{\!p}{=}1$, $\beta_{\!h}{=}1$). Optimized on $n{=}50$ with 5 instances.

    \item \textbf{Bin Packing with ACO framework (BPP-ACO):} Given $n$ items with demands $d_i \sim \text{Uniform}[20, 100]$ and bin capacity $C{=}150$, minimize the number of bins $B$ such that $\sum_{i \in B_j} d_i \leq C$ for all $j$ \cite{ross2002hyper}. The LLM generates an $(n \times n)$ heuristic matrix from the demand vector. ACO parameters: 20 ants, 15 iterations, $\rho{=}0.9$, $\alpha_{\!p}{=}1$, $\beta_{\!h}{=}1$. Optimized on $n{=}500$ with 5 instances.

    \item \textbf{Adaptive Curriculum Sequencing (ACS):} A real-world problem from e-learning: generate personalized learning paths balancing learner characteristics with pedagogical constraints. Using the OULAD dataset \cite{martins2021comparative} with 120 materials, 20 concepts, and 30 students, the decision variable is $X \in \{0,1\}^{30 \times 120}$. The objective function minimizes concept coverage gaps, time-bound violations, and preference mismatches:
    \begin{multline}
        f(X) = \varepsilon_1 \omega_1 \sum_i \max(0, \text{covered}_i - |\mathbf{g}_i|) \\
        + \varepsilon_2 \omega_2 \sum_i (1 - \text{coverage}_i) + \varepsilon_3 \sum_i \mathbf{1}\{\text{time}_i \notin [t_i^{\min}, t_i^{\max}]\} \\
        + \varepsilon_4 \sum_i \text{pref\_mismatch}_i
    \end{multline}
    with $\varepsilon_1{=}1$, $\varepsilon_2{=}10^4$, $\varepsilon_3{=}1000$, $\omega_1{=}0.25$, $\omega_2{=}1$, $\varepsilon_4{=}0.25$, where $[t_i^{\min}, t_i^{\max}]$ is the student-specific attention span range, and material priority limits are $\psi_1 = 3$ (high), $\psi_2 = 6$ (medium), $\psi_3 = 1$ (challenging). A guiding radius $rg$ decays linearly from 2.5 to 0.1. Evaluation: 20 search agents, 50 iterations. See \textbf{TABLE~\ref{tab:constraint_params}}.

    \item \textbf{Wireless Sensor Network (WSN) Deployment:} Optimize placement and power of 50 Convergence Nodes (CNs) to cover 200 fixed Sensing Nodes (SNs) while minimizing energy. Each CN $j$ has position $(x_j, y_j) \in [0,50]^2$ and power $P_j \in [0,30]$ dBm. Path loss: $PL_{ij} = 55.0 + 10 \times 2.5 \times \log_{10}(d_{ij}) + \beta_{ij}$, where $\beta_{ij} = \beta^{x}_{ij} + \beta^{y}_{ij}$ is a position-dependent environmental fading: $\beta^{x}_{ij}{=}8.0$ (resp.\ $\beta^{y}_{ij}{=}20.0$) when SN $i$ and CN $j$ lie on opposite sides of the $x{=}25$ (resp.\ $y{=}25$) midline, and 0 otherwise. Constraints: received signal $\geq -85$ dBm, CN connectivity (range 20.0), capacity $\leq 15$ SNs per CN. Objective:
    \begin{multline}
        f(\mathbf{x}) = \sum_j 10^{P_j/10} + 10 \times \text{uncovered\_SNs} \\
        + 1000 \times \text{disconnected} + 100 \times \max(0, \text{std}(P) - 1).
    \end{multline}
    Guiding radius $rg$ decays from 2.5 to 0.1. Evaluation: 50 search agents, 100 iterations. See \textbf{TABLE~\ref{tab:constraint_params}}.

    \item \textbf{Traveling Salesperson Problem, Constructive (TSP-Construct):} Given $n$ city coordinates $\{c_1, \ldots, c_n\} \subset \mathbb{R}^2$ normalized to $[0,1]$, find the shortest tour visiting all cities exactly once: $\min \sum_{i=1}^{n-1} d(\pi_i, \pi_{i+1}) + d(\pi_n, \pi_1)$, where $\pi$ is a permutation and $d(\cdot,\cdot)$ is the Euclidean distance. The LLM generates a \emph{constructive heuristic}, a function called iteratively to select the next city at each step. To provide spatial locality information, we introduce a neighborhood matrix $N \in \mathbb{Z}^{n \times n}$ where $N_i$ lists nodes sorted by distance from node $i$. At each step, candidates are drawn from the neighborhood of the current node:
    \begin{equation}
        \tilde{U}_t = \{j \in N_{\pi_t} : j \notin \{\pi_1, \ldots, \pi_t\}\}.
    \end{equation}
    Training: 64 instances with $n{=}50$, evaluation budget $L{=}100$. Evaluation: $n \in \{50, 100, 200\}$, 64 instances each.
\end{itemize}

\textbf{TABLE~\ref{tab:benchmark_params}} summarizes the parameter settings for all five optimization problems. Settings for TSP-ACO, BPP-ACO and TSP-Construct follow previous studies \cite{ye2024reevo,zheng2025monte}; ACS and WSN parameters were determined through extensive testing.

\begin{table}[htbp]
\centering
\resizebox{\columnwidth}{!}{%
\begin{tabular}{cccccc}
\toprule
\textbf{Parameter} & \textbf{TSP-ACO} & \textbf{BPP-ACO} & \textbf{ACS} & \textbf{WSN} & \textbf{TSP-Construct} \\
\midrule
Initial Population $N_0$ & 20 & 20 & 20 & 20 & 20 \\
Offspring per Gen.\ $N_\mathrm{pop}$ & 10 & 10 & 10 & 10 & 10 \\
Max Solutions $L$ & 100 & 50 & 50 & 50 & 100 \\
Independent Runs & 5 & 5 & 5 & 5 & 5 \\
\addlinespace
\midrule
\multicolumn{6}{c}{\textbf{Problem-specific}} \\
\midrule
Number of items/cities $n$ & 50 & 500 & -- & -- & 50 / 100 / 200 \\
Item demand $d_i$ & -- & $\text{Uniform}[20,100]$ & -- & -- & -- \\
Bin capacity $C$ & -- & 150 & -- & -- & -- \\
\addlinespace
Materials & -- & -- & 120 & -- & -- \\
Concepts & -- & -- & 20 & -- & -- \\
Students & -- & -- & 30 & -- & -- \\
Number of CN & -- & -- & -- & 50 & -- \\
Number of SN & -- & -- & -- & 200 & -- \\
Capacity & -- & -- & -- & 15 & -- \\
Connection Distance & -- & -- & -- & 20 & -- \\
SN Connect Param.\ & -- & -- & -- & $-85$ & -- \\
Path Loss Const.\ $PL_0$ & -- & -- & -- & 55.0 & -- \\
$\gamma_{\!PL}$ (Path loss exponent) & -- & -- & -- & 2.5 & -- \\
\addlinespace
\midrule
Search Agents & -- & -- & 20$^{\dagger}$ & 50$^{\dagger}$ & -- \\
Max Iterations & -- & -- & 50$^{\dagger}$ & 100$^{\dagger}$ & -- \\
Initial $rg$ & -- & -- & 2.5 & 2.5 & -- \\
Min $rg$ & -- & -- & 0.1 & 0.1 & -- \\
\bottomrule
\end{tabular}%
}
\vspace{2pt}
\par\raggedright\footnotesize
$^{\dagger}$For ACS and WSN, traditional baselines (ACO, GA, PSO, SCSO) use increased budgets (ACS: 30 agents, 500 iterations; WSN: 100 agents, 1000 iterations), also independently run 5 times.
\caption{Parameter settings for all five optimization problems.}
\label{tab:benchmark_params}
\end{table}

\textbf{TABLE~\ref{tab:constraint_params}} lists the penalty coefficients and constraint parameters specific to the ACS and WSN problems.

\begin{table}[htbp]
\centering
\begin{tabular}{@{}c >{\centering\arraybackslash}p{5cm}c@{}}
\toprule
Parameter & Description & Value \\
\midrule
\multicolumn{3}{c}{\textbf{ACS}} \\
\midrule
$\varepsilon_{1}$ & Penalty factor for exceeding the number of concepts required for learning & 1 \\
$\varepsilon_{2}$ & Penalty factor for not covering the number of learning concepts & $10^4$ \\
$\varepsilon_{3}$ & Penalty per time-bound violation (outside $[t_i^{\min}, t_i^{\max}]$) & 1000 \\
$\omega_{1}$ & Coefficient for $\varepsilon_{1}$ & 0.25 \\
$\omega_{2}$ & Coefficient for $\varepsilon_{2}$ & 1 \\
$\varepsilon_{4}$ & Penalty factor for preference mismatch & 0.25 \\
$\psi_{1}$ & High priority material quantity limit & 3 \\
$\psi_{2}$ & Medium priority material quantity limit & 6 \\
$\psi_{3}$ & Challenging material quantity limit & 1 \\
\midrule
\multicolumn{3}{c}{\textbf{WSN}} \\
\midrule
coverage & Penalty per uncovered sensor node & 10 \\
connectivity & Penalty for disconnected network & 1000 \\
power\_std & Penalty for power standard deviation & 100 \\
$\beta^{x}$ & Cross-midline $x$-fading (opposite sides of $x{=}25$) & 8.0 \\
$\beta^{y}$ & Cross-midline $y$-fading (opposite sides of $y{=}25$) & 20.0 \\
\bottomrule
\end{tabular}%
\caption{Constraint parameters for ACS and WSN problems.}
\label{tab:constraint_params}
\end{table}

\textbf{TABLE~\ref{tab:meta_params}} details the parameters of all baseline methods and AHD architectures.

\begin{table}[htbp]
\centering
\small
\resizebox{\columnwidth}{!}{%
\begin{tabular}{@{}c >{\centering\arraybackslash}p{7.2cm}@{}}
\toprule
\textbf{Method} & \textbf{Parameters} \\
\midrule
\multicolumn{2}{c}{\textit{Baseline Methods}} \\
\midrule
Greedy Construct & -- \\ \addlinespace
GP & $\mu{=}1000$, $G{=}200$, $r_c{=}0.9$, $r_m{=}0.1$, $k_{tour}{=}7$, $D_{\max}{=}17$, $d_{\max}{=}6$, $e{=}1$, $\gamma{=}0.005$ \\ \addlinespace
ACO (TSP-ACO) & $n_{ants}=30$, $n_{iter}=100$, $\alpha_{\!p}=1$, $\beta_{\!h}=1$, $\rho=0.9$ \\ \addlinespace
ACO (BPP-ACO) & $n_{ants}=20$, $n_{iter}=15$, $\alpha_{\!p}=1$, $\beta_{\!h}=1$, $\rho=0.9$ \\ \addlinespace
SCSO & $rG$ decays from $2$ to $0$ \\ \addlinespace
GA & crossover $=0.8$, mutation $=0.1$ \\ \addlinespace
PSO & $w=0.7$, $c_1{=}c_2{=}1.5$ \\
\midrule
\multicolumn{2}{c}{\textit{AHD Architecture}} \\
\midrule
Funsearch & $k=2$ \\ \addlinespace
EoH & $p=5$ \\ \addlinespace
ReEvo & crossover rate $=1$, mutation rate $=0.5$ \\ \addlinespace
MCTS-AHD & $NI=4$, $H=10$, $k=2$, $p_{0}\in\{2,3,4,5\}$, $\lambda_0=0.1$, $\alpha_{MCTS}=0.5$ \\ \addlinespace
HiFo-Prompt & Multiple parameters; see \cite{chen2025hifo} \\ \addlinespace
MeLA & $COR = 2$ \\ \addlinespace
\textbf{MeEvo} & $COR = 2$, $\alpha=0.5$, $\beta=0.2$, $K=5$, $N=2$, $M=1$ \\
\bottomrule
\end{tabular}%
}
\caption{Parameters of Baseline Methods and AHD Architectures}
\label{tab:meta_params}
\end{table}

\textbf{TABLE~\ref{tab:llm_config}} shows the LLM backbone configurations. All five problems are evaluated on both backbones.

\begin{table}[htbp]
\centering
\resizebox{\columnwidth}{!}{%
\begin{tabular}{ccc}
\toprule
\textbf{LLM Model} & \textbf{Temperature} & \textbf{Description} \\
\midrule
DeepSeek-V4-Flash & 1.0 & High exploration, diverse generation \\
MIMO-v2.5-Pro & 0.5 & Lower randomness, more stable code \\
\bottomrule
\end{tabular}%
}
\caption{LLM backbone configurations. All five problems are evaluated on both backbones.}
\label{tab:llm_config}
\end{table}

Each problem is independently run \textbf{5} times. For TSP-Construct, we report the mean (standard deviations are small across architectures). For the remaining problems, we report mean, standard deviation, and best value following the protocol of prior work \cite{ye2024reevo}. All problems are minimization tasks (lower is better).

\textbf{TSP-ACO, BPP-ACO, ACS, and WSN} (\textbf{TABLE~\ref{tab:comparative}}) follow an indirect protocol: the LLM generates heuristic information that guides a downstream solver (ACO) or directly optimizes solution components. We compare against ACO \cite{dorigo2006ant}, GA \cite{holland1992genetic}, PSO \cite{kennedy1995particle}, and SCSO \cite{seyyedabbasi2023sand} as traditional baselines, plus LLM architectures. For ACS and WSN, we increase traditional baselines' computational budgets following prior metaheuristic studies on these problems \cite{martins2021comparative,machado2021metaheuristic,chen2024optimizing}. Traditional baselines are thus evaluated with 30 agents and 500 iterations for ACS, and 100 agents and 1000 iterations for WSN, substantially exceeding the LLM-based AHD protocol (ACS: 20 agents, 50 iterations; WSN: 50 agents, 100 iterations). This asymmetry reflects the distinct search paradigms: traditional metaheuristics start from random initialization and rely on large evaluation budgets to discover structure, whereas LLM-based architectures encode problem knowledge into heuristic design before evaluation begins. Each traditional baseline is also independently executed 5 times, following the same protocol as the LLM-based AHD architectures (\textbf{TABLE~\ref{tab:benchmark_params}}).

\textbf{TSP-Construct} evaluates constructive heuristics. Optimal tours are from the LKH solver \cite{lin1973effective,zheng2025monte}. We compare against Nearest-greedy \cite{rosenkrantz1977analysis}, POMO \cite{kwon2020pomo}, GP \cite{koza1994genetic}, and LLM architectures (FunSearch \cite{romera2024mathematical}, EoH \cite{liu2024evolution}, ReEvo \cite{ye2024reevo}, MCTS-AHD \cite{zheng2025monte}, HiFo-Prompt \cite{chen2025hifo}, MeLA \cite{QIU2026133022}). Gap is relative to LKH optimal.

For Genetic Programming (GP) in TSP-Construct, we implement tree-based genetic programming with ramped half-and-half initialization following Koza~\cite{koza1994genetic,poli2007genetic}; all GP parameters are listed in \textbf{TABLE~\ref{tab:meta_params}}. Briefly, a population of $\mu{=}1000$ expression trees is evolved over $G{=}200$ generations, using tournament selection ($k_{tour}{=}7$), one-point subtree crossover ($r_c{=}0.9$), and uniform subtree mutation ($r_m{=}0.1$)~\cite{poli2007genetic}. Tree depth is bounded by $D_{\max}{=}17$ during evolution and $d_{\max}{=}6$ at initialization to curb bloat~\cite{koza1994genetic}; one elite individual is preserved per generation ($e{=}1$), and a parsimony coefficient $\gamma{=}0.005$ penalizes tree size in the fitness function to favor compact expressions. The function set comprises $\{+,-,\times,\div,\max,\min,\sin,\cos,\sqrt{\cdot},\text{neg},\text{abs}\}$ with ephemeral random constants in $[-5,5]$; the terminal set supplies eight real-valued TSP heuristic features (e.g., current-to-candidate distance, unvisited-city statistics, normalized path length). The population is evolved on 10 randomly sampled $n{=}50$ TSP training instances; the best-evolved individual is evaluated on 64 held-out test instances at each of $n{\in}\{50, 100, 200\}$.

POMO, GP, and HiFo-Prompt are evaluated only on TSP-Construct. POMO~\cite{kwon2020pomo} is a deep RL model trained end-to-end for TSP city selection and cannot generate the heuristic matrices (TSP-ACO, BPP-ACO) or continuous search operators (ACS, WSN) required by the other four problems. GP~\cite{koza1994genetic} evolves symbolic expression trees designed for scalar constructive heuristics; adapting to matrix-valued or continuous-space outputs demands a fundamentally different function set. HiFo-Prompt~\cite{chen2025hifo} requires extensive per-problem manual parameter configuration (\textbf{TABLE~\ref{tab:meta_params}}), rendering cross-problem deployment impractical.

Parameters are detailed in \textbf{TABLE~\ref{tab:meta_params}}. The rationale for optimization parameters ($\alpha$, $\beta$, $K$) is provided in the sensitivity analysis below. $N$ and $M$ define the cyclic alternation, the core architectural contribution of MeEvo, and are analyzed in the ablation study (\S\ref{subsec:ablation}).

\subsection{Parameter Sensitivity Analysis}
\label{subsec:param_sensitivity}

We distinguish two parameter categories. $\alpha$, $\beta$, and $K$ are optimization hyperparameters internal to Natural Evolution; their sensitivity is examined below. $N$ and $M$ govern the cyclic alternation and are ablated in \S\ref{subsec:ablation} to validate the architectural design. We evaluate variants on two representative problems, TSP-ACO (constrained combinatorial) and ACS (real-world, multi-constraint), using DeepSeek-V4-Flash, with all other parameters held at their default values (\textbf{TABLE~\ref{tab:meta_params}}). Each configuration is repeated 5 times; we report the mean objective.

\textbf{TABLE~\ref{tab:sensitivity}} reports the sensitivity of all three optimization hyperparameters. Panel (a) examines $\alpha$ and $\beta$. The default configuration ($\alpha{=}0.5$, $\beta{=}0.2$) is compared against a fixed-probability baseline ($\beta{=}0$), wider amplitude ($\beta{=}0.4$), narrower amplitude ($\beta{=}0.1$), a higher baseline ($\alpha{=}0.6$, $\beta{=}0.1$), and a reversed direction ($\alpha{=}0.7$, $\beta{=}-0.2$, i.e., $p_c$ increases over time). Panel (b) varies $K$ which controls the selection pressure in Natural Evolution. The default ($K{=}5$) is compared against smaller pools ($K \in \{2, 3\}$) and larger pools ($K \in \{8, 10\}$).

\begin{table}[htbp]
\centering
\small
\caption{Sensitivity analysis of optimization hyperparameters.}
\label{tab:sensitivity}
\resizebox{\columnwidth}{!}{%
\begin{tabular}{ccccc}
\toprule
\multicolumn{5}{c}{\textbf{Panel A: Adaptive operator balance hyperparameters $\alpha$ and $\beta$}} \\
\midrule
$(\alpha, \beta)$ & $p_c$ Range & TSP-ACO Obj. ($\downarrow$) & ACS Obj. ($\downarrow$) & Notes \\
\midrule
$(0.5, 0.0)$ & [0.5, 0.5] & 5.736 & 584.74 & Fixed (no adaptation) \\
$(0.5, 0.1)$ & [0.5, 0.6] & 5.728 & 579.42 & Narrow amplitude \\
$(0.5, 0.4)$ & [0.5, 0.9] & 5.730 & 579.12 & Wide amplitude \\
$(0.6, 0.1)$ & [0.6, 0.7] & 5.742 & 590.73 & Higher baseline \\
$(0.7, -0.2)$ & [0.5, 0.7] & 5.724 & 582.18 & Reversed ($p_c \uparrow$) \\
\midrule
$(0.5, \mathbf{0.2})$ & [0.5, 0.7] & \textbf{5.716} & \textbf{578.16} & \textbf{Default} \\
\midrule
\multicolumn{5}{c}{\textbf{Panel B: Parent pool size $K$}} \\
\midrule
$K$ & --- & TSP-ACO Obj. ($\downarrow$) & ACS Obj. ($\downarrow$) & Notes \\
\midrule
2 & --- & 5.772 & 591.83 & Minimum viable \\
3 & --- & 5.754 & 584.15 & Small pool \\
8 & --- & 5.724 & 582.83 & Large pool \\
10 & --- & 5.793 & 602.43 & Very large pool \\
\midrule
$\mathbf{5}$ & --- & \textbf{5.716} & \textbf{578.16} & \textbf{Default} \\
\bottomrule
\end{tabular}%
}
\end{table}

The results in \textbf{TABLE~\ref{tab:sensitivity}} support three conclusions regarding the operator balance. First, the linear shift consistently outperforms a fixed crossover probability ($\beta{=}0$), confirming that shifting from exploration-dominated (crossover) to exploitation-balanced (mutation) over the course of search is beneficial. Second, the amplitude $\beta{=}0.2$ strikes the right balance: a narrower amplitude ($\beta{=}0.1$) provides too little shift, while a wider amplitude ($\beta{=}0.4$) over-oscillates and degrades performance on ACS. Third, the baseline $\alpha{=}0.5$ is critical: raising it to $0.6$ sharply degrades ACS performance (from 578.16 to 590.73), indicating that mutation pressure must reach approximately 50\% by convergence to enable sufficient local refinement. The reversed direction ($\alpha{=}0.7$, $\beta{=}-0.2$) underperforms the default, confirming that exploration should dominate early and gradually give way to exploitation, not the reverse. These findings align with the meso-level design principle in \S\ref{subsec:ee_review}: crossover (exploration) should be favored at the start of search, with mutation (exploitation) progressively increasing as the population converges.

Regarding parent pool size (Panel B), \textbf{TABLE~\ref{tab:sensitivity}} reveals a U-shaped sensitivity. $K{=}2$ underperforms due to insufficient parent diversity for meaningful crossover (TSP rises to 5.772, ACS to 591.83). $K{=}10$ performs worst overall (TSP: 5.793, ACS: 602.43), as a large pool admits lower-fitness individuals that dilute selection pressure and steer the LLM toward weaker design patterns. $K{=}5$ achieves the optimal trade-off: it maintains sufficient diversity for effective recombination without relaxing selection pressure. This is consistent with truncation selection theory in evolutionary computation, where moderate selection pressure balances exploration and exploitation.

\subsection{Results and Analysis}

\textbf{TSP-ACO, BPP-ACO, ACS, and WSN.} \textbf{TABLE~\ref{tab:comparative}} presents results on four problems. MeEvo achieves the best mean performance across all problems with both LLM backbones, with gaps to the next best architecture below 1.68\% on TSP-ACO and 0.25\% on BPP-ACO. On ACS and WSN, the differences are larger. Despite receiving substantially greater computational budgets, traditional baselines struggle: even SCSO, the best traditional method, reaches Gaps of 189.46\% (ACS) and 720.91\% (WSN). This is consistent with the No Free Lunch theorem \cite{wolpert1997no}: traditional metaheuristics are problem-agnostic algorithms that apply the same search operators regardless of problem structure, whereas LLM-based architectures generate problem-specific heuristics that encode domain knowledge into the search strategy. On complex constrained problems such as ACS and WSN, this problem-specificity provides a substantial advantage. Among LLM-based architectures, MeEvo outperforms the next best MeLA on both backbones (e.g., MeLA Gap: 1.79\% on ACS, 14.96\% on WSN with DeepSeek-V4-Flash), and shows superior stability ($\pm5.21$ on ACS, $\pm0.76$ on WSN). Best heuristic codes are provided in the Appendix.

\begin{table*}
\centering
\caption{Comparative Performance Analysis on TSP-ACO, BPP-ACO and real-world problems. For ACS and WSN, traditional baselines are evaluated with increased resources (ACS: 30 search agents, 500 iterations; WSN: 100 search agents, 1000 iterations) to ensure fair comparison with LLM-based methods. All results are based on 5 independent runs.}
\label{tab:comparative}
\begin{tabular}{ccccccccc}
\toprule
\multirow{2}{*}{Method} & \multicolumn{2}{c}{TSP-ACO} & \multicolumn{2}{c}{BPP-ACO} & \multicolumn{2}{c}{ACS} & \multicolumn{2}{c}{WSN} \\
\cmidrule(lr){2-3} \cmidrule(lr){4-5} \cmidrule(lr){6-7} \cmidrule(lr){8-9}
 & Obj ($\downarrow$) & Gap (\%) & Obj  ($\downarrow$) & Gap (\%) & Obj  ($\downarrow$) & Gap (\%) & Obj ($\downarrow$) & Gap (\%) \\
\midrule
ACO & $5.98 \pm 0.19$ & 4.62 & $208.54 \pm 1.37$ & 2.09 & $1793.72 \pm 108.32$ & 210.25 & $1294.70 \pm 80.32$ & 2448.62 \\
GA & - & - & - & - & $2634.25 \pm 218.92$ & 355.63 & $1922.46 \pm 0.13$ & 3684.37 \\
PSO & - & - & - & - & $2662.31 \pm 80.34$ & 360.48 & $1800.28 \pm 286.34$ & 3443.86 \\
SCSO & - & - & - & - & $1673.53 \pm 112.04$ & 189.46 & $417.02 \pm 9.88$ & 720.91 \\
\midrule
\multicolumn{9}{c}{LLM backbone: \textit{DeepSeek-V4-Flash}} \\
\midrule
Funsearch & $5.923 \pm 0.02$ & 3.62 & $206.60 \pm 0.15$ & 1.14 & $1242.75 \pm 639.42$ & 114.94 & $96.38 \pm 27.32$ & 89.72 \\
EoH & $5.852 \pm 0.01$ & 2.38 & $205.48 \pm 0.75$ & 0.59 & $1253.97 \pm 462.30$ & 116.89 & $119.71 \pm 15.77$ & 135.65 \\
ReEvo & $5.833 \pm 0.01$ & 2.05 & $205.48 \pm 0.97$ & 0.59 & $654.95 \pm 22.85$ & 13.28 & $130.06 \pm 24.47$ & 156.02 \\
MCTS-AHD & $5.812 \pm 0.02$ & 1.68 & $205.28 \pm 0.72$ & 0.49 & $1213.48 \pm 434.93$ & 109.89 & $83.36 \pm 6.83$ & 64.09 \\
MeLA & $5.828 \pm 0.01$ & 1.96 & $204.80 \pm 0.28$ & 0.25 & $588.50 \pm 9.07$ & 1.79 & $58.40 \pm 2.02$ & 14.96 \\
\midrule
\textbf{MeEvo} & \textbf{\boldmath$5.716 \pm 0.00$} & \textbf{0.00} & \textbf{\boldmath$204.28 \pm 0.10$} & \textbf{0.00} & \textbf{\boldmath$578.16 \pm 5.21$} & \textbf{0.00} & \textbf{\boldmath$50.80 \pm 0.76$} & \textbf{0.00} \\
\midrule
\multicolumn{9}{c}{LLM backbone: \textit{MIMO-v2.5-Pro}} \\
\midrule
Funsearch & $6.013 \pm 0.12$ & 5.20 & $207.20 \pm 1.33$ & 1.43 & $687.24 \pm 34.78$ & 18.87 & $99.23 \pm 45.71$ & 95.33 \\
EoH & $5.847 \pm 0.03$ & 2.29 & $204.88 \pm 0.43$ & 0.29 & $1273.18 \pm 525.10$ & 120.21 & $104.89 \pm 10.43$ & 106.48 \\
ReEvo & $5.838 \pm 0.02$ & 2.13 & $204.68 \pm 0.52$ & 0.20 & $1239.17 \pm 592.81$ & 114.33 & $125.94 \pm 31.69$ & 147.91 \\
MCTS-AHD & $5.807 \pm 0.01$ & 1.59 & $204.44 \pm 0.04$ & 0.08 & $1392.67 \pm 519.74$ & 140.88 & $92.82 \pm 5.34$ & 82.72 \\
MeLA & $5.824 \pm 0.01$ & 1.89 & $204.60 \pm 0.20$ & 0.16 & $609.47 \pm 15.21$ & 5.42 & $59.37 \pm 3.16$ & 16.87 \\
\midrule
\textbf{MeEvo} & \textbf{\boldmath$5.718 \pm 0.01$} & \textbf{0.04} & \textbf{\boldmath$204.40 \pm 0.24$} & \textbf{0.06} & \textbf{\boldmath$585.94 \pm 11.14$} & \textbf{1.35} & \textbf{\boldmath$52.00 \pm 2.26$} & \textbf{2.36} \\
\bottomrule
\end{tabular}%
\end{table*}

\begin{figure*}[t]
    \centering
    \includegraphics[width=\textwidth]{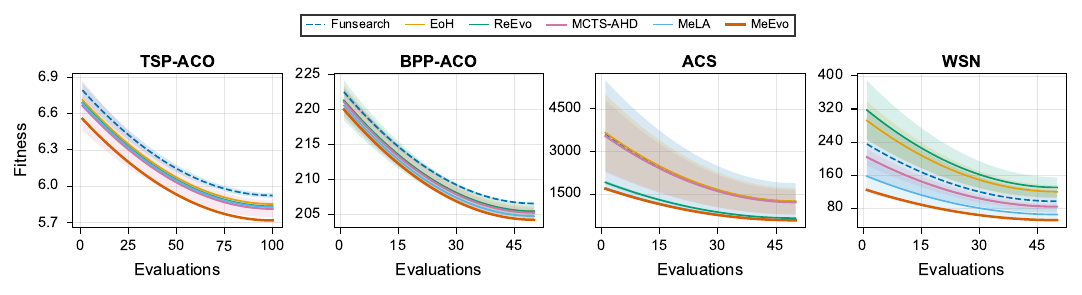}
    \caption{Convergence curves. On TSP-ACO and BPP-ACO, all architectures converge to similar values. On ACS and WSN, MeEvo achieves lower final objectives.}
    \label{fig:fitness}
\end{figure*}

\textbf{FIG.~\ref{fig:fitness}} shows convergence curves. On TSP-ACO and BPP-ACO, all architectures converge, with MeEvo achieving marginally lower values. On ACS and WSN, FunSearch and MCTS-AHD exhibit erratic convergence with large fluctuations, while MeEvo maintains smooth descent, converges to better solutions.

\begin{figure*}[t]
    \centering
    \includegraphics[width=\textwidth]{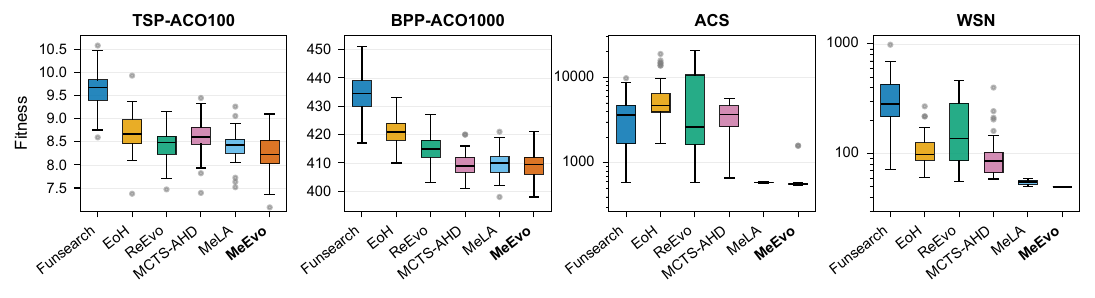}
    \caption{Performance of Optimal Heuristic Generated by Different Architectures.}
    \label{fig:sta}
\end{figure*}

LLM-based architectures converge within far fewer evaluations than traditional baselines, raising a further question: is the best heuristic found by chance, or does it reflect a genuinely effective solution that generalizes and remains reproducible? \textbf{FIG.~\ref{fig:sta}} tests the best heuristic from each architecture (\textbf{TABLE~\ref{tab:best_values}}): for TSP-ACO (trained at $n{=}50$) and BPP-ACO (trained at $n{=}500$), it is evaluated on 64 instances at $n{=}100$ and $n{=}1000$; for ACS and WSN, it is independently executed 30 times. In TSP-ACO, the best heuristic generated by MeEvo exhibits the lowest median value. In ACS and WSN scenarios, the best heuristic generated by MeEvo demonstrated the stable and superior performance, where the heuristic not only exhibited minimal deviation but consistently converged to its optimal fitness value in nearly every run, confirming its exceptional quality and reliability. These results suggest that the best heuristics produced by MeEvo generalize to larger scales and maintain reasonable reproducibility across independent runs.

\begin{table}[htbp]
\centering
\small
\resizebox{\columnwidth}{!}{%
\begin{tabular}{ccccc}
\toprule
\multirow{2}{*}{Method} & TSP-ACO & BPP-ACO & ACS & WSN \\
\cmidrule(lr){2-2} \cmidrule(lr){3-3} \cmidrule(lr){4-4} \cmidrule(lr){5-5}
 & Best Obj. & Best Obj. & Best Obj. & Best Obj. \\
\midrule
Funsearch & $5.902$ & $206.00$ & $652.27$ & $76.19$ \\
EoH & $5.842$ & $204.60$ & $642.75$ & $98.38$ \\
ReEvo & $5.823$ & $\mathbf{204.20}$ & $628.06$ & $94.61$ \\
MCTS-AHD & $5.792$ & $204.40$ & $642.50$ & $76.10$ \\
MeLA & $5.812$ & $204.40$ & $579.32$ & $55.76$ \\
\midrule
\textbf{MeEvo (DeepSeek)} & $\mathbf{5.712}$ & $\mathbf{204.20}$ & $\mathbf{568.38}$ & $\mathbf{50.00}$ \\
\textbf{MeEvo (MIMO)} & $\mathbf{5.712}$ & $\mathbf{204.20}$ & $\mathbf{564.44}$ & $\mathbf{50.00}$ \\
\bottomrule
\end{tabular}%
}
\caption{Best values across different architectures.}
\label{tab:best_values}
\end{table}

\textbf{TSP-Construct.} \textbf{TABLE~\ref{tab:tsp_construct}} compares constructive heuristics across three problem sizes. POMO achieves near-optimal performance at $n{=}50$ (Gap 0.32\%) through deep reinforcement learning trained on thousands of $n{=}50$ instances, but collapses to 20.80\% at $n{=}200$. GP achieves Gaps of 11.17\%, 12.63\%, and 13.85\%, surpassing all LLM-based AHD architectures at $n{=}50$ and $n{=}200$, with far stronger generalization (11.17\%$\rightarrow$13.85\%) than POMO (0.32\%$\rightarrow$20.80\%). However, GP incurs approximately 2 million TSP evaluations during evolution ($\mu{=}1000 \times G{=}200 \times 10$ training instances), compared to the zero-shot or few-shot LLM-based architectures. These three paradigms occupy distinct points on a trade-off spectrum: POMO trades generality for specialization, GP trades evaluation cost for structural generalization, and LLM-based AHD trades evaluation density for cross-domain applicability. Notably, LLM-based methods lag behind GP and POMO on this task. In TSP-ACO and BPP-ACO, the LLM only needs to encode search guidance (a heuristic matrix) while ACO provides iterative optimization that compensates for suboptimal heuristics. In TSP-Construct, the LLM must directly generate a complete constructive procedure with no downstream solver; it cannot learn from evaluation feedback the way ACO does, and each design decision is committed without revision. Under DeepSeek-V4-Flash, MeEvo ranks first among LLM-based methods at all three sizes (Gaps of 11.55\%, 11.84\%, and 14.59\%). Under MIMO-v2.5-Pro, it ranks first at $n{=}50$ (11.71\%) and $n{=}200$ (14.55\%), and third at $n{=}100$.

\begin{table}
\centering
\caption{Comparative Performance Analysis on TSP-Construct Across Different Problem Sizes}
\label{tab:tsp_construct}
\resizebox{\columnwidth}{!}{%
\begin{tabular}{ccccccc}
\toprule
\multirow{2}{*}{Method} & \multicolumn{2}{c}{$n=50$} & \multicolumn{2}{c}{$n=100$} & \multicolumn{2}{c}{$n=200$} \\
\cmidrule(lr){2-3} \cmidrule(lr){4-5} \cmidrule(lr){6-7}
 & Obj ($\downarrow$) & Gap (\%) & Obj ($\downarrow$) & Gap (\%) & Obj ($\downarrow$) & Gap (\%) \\
\midrule
Optimal  & 5.678 & - & 7.767 & - & 10.658 & - \\
Greedy Construct & 7.003 & 23.34 & 9.542 & 22.85 & 13.094 & 22.86 \\
POMO & \textbf{5.696} & \textbf{0.32} & \textbf{8.002} & \textbf{3.03} & 12.875 & 20.80 \\
GP & 6.312 & 11.17 & 8.748 & 12.63 & \textbf{12.134} & \textbf{13.85} \\
\midrule
\multicolumn{7}{c}{LLM backbone: \textit{DeepSeek-V4-Flash}} \\
\midrule
Funsearch & 6.803 & 19.81 & 9.182 & 18.22 & 12.850 & 20.57 \\
EoH & 6.485 & 14.21 & 8.894 & 14.51 & 12.582 & 18.05 \\
ReEvo & 6.622 & 16.63 & 8.893 & 14.50 & 12.362 & 15.99 \\
MCTS-AHD & 6.403 & 12.77 & 8.762 & 12.81 & 12.271 & 15.13 \\
HiFo-Prompt & 6.392 & 12.57 & 8.749 & 12.64 & 12.264 & 15.07 \\
MeLA & 6.423 & 13.12 & 8.801 & 13.31 & 12.338 & 15.76 \\
MeEvo(ours) & \cellcolor{gray!15}6.334 & \cellcolor{gray!15}11.55 & \cellcolor{gray!15}8.687 & \cellcolor{gray!15}11.84 & \cellcolor{gray!15}12.213 & \cellcolor{gray!15}14.59 \\
\midrule
\multicolumn{7}{c}{LLM backbone: \textit{MIMO-v2.5-Pro}} \\
\midrule
Funsearch & 6.927 & 22.00 & 9.034 & 16.31 & 12.737 & 19.51 \\
EoH & 6.473 & 14.00 & 8.832 & 13.71 & 12.682 & 18.99 \\
ReEvo & 6.732 & 18.56 & 8.816 & 13.51 & 12.536 & 17.62 \\
MCTS-AHD & 6.403 & 12.77 & 8.697 & 11.97 & 12.290 & 15.31 \\
HiFo-Prompt & 6.392 & 12.57 & \cellcolor{gray!15}8.693 & \cellcolor{gray!15}11.92 & 12.285 & 15.27 \\
MeLA & 6.423 & 13.12 & 8.795 & 13.24 & 12.413 & 16.47 \\
MeEvo(ours) & \cellcolor{gray!15}6.343 & \cellcolor{gray!15}11.71 & 8.753 & 12.69 & \cellcolor{gray!15}12.209 & \cellcolor{gray!15}14.55 \\
\bottomrule
\end{tabular}%
}
\end{table}

\subsection{Ablation Analysis}
\label{subsec:ablation}

MeEvo alternates between $N$ generations of Natural Evolution and $M$ generations of Metacognitive Evolution. To test whether both processes are necessary and whether the alternation ratio matters, we evaluate four ablated configurations with DeepSeek-V4-Flash (5 runs each): (i) Natural Evolution only, denoted $(N, /)$; (ii) Metacognitive Evolution only, denoted $(/, M)$; (iii) balanced alternation $(1,1)$; and (iv) reflection-heavy alternation $(1,2)$. \textbf{TABLE~\ref{tab:ablation_nm}} summarizes the full results.

\textbf{Single-process baselines.} Removing Metacognitive Evolution $(N, /)$ collapses on ACS (1293.54) and WSN (98.19), confirming that population-driven exploration alone cannot handle constrained tasks; on TSP-ACO it ties with $(/, M)$ (5.83), as the ACO framework supplies sufficient structure. $(/, M)$ performs substantially better on ACS (588.50) and WSN (58.40), showing metacognitive guidance is essential, but is outperformed by all combined configurations, confirming that population diversity provides complementary benefit.

\textbf{Combined process configurations.} Both $(1, 1)$ and $(1, 2)$ outperform the single-process baselines across all problems, confirming that even minimal alternation helps. $(1, 1)$ and $(1, 2)$ perform comparably overall: $(1, 1)$ leads on TSP-ACO (5.75), BPP-ACO (204.44), and WSN (53.37), while $(1, 2)$ leads on ACS (581.28). The default $(2, 1)$ clearly outperforms both (TSP-ACO 5.72, BPP-ACO 204.28, ACS 578.16, WSN 50.80), confirming that sufficient Natural Evolution exploration combined with unified Metacognitive reflection yields the best performance.

\begin{table}[htbp]
\centering
\small
\resizebox{\columnwidth}{!}{%
\begin{tabular}{ccccc}
\toprule
$(N, M)$ & TSP-ACO & BPP-ACO & ACS & WSN \\
 & Obj. ($\downarrow$) & Obj. ($\downarrow$) & Obj. ($\downarrow$) & Obj. ($\downarrow$) \\
\midrule
$(N, /)$ -- Natural Evo & 5.83 & 205.28 & 1293.54 & 98.19 \\
$(/, M)$ -- Metacognitive Evo & 5.83 & 204.80 & 588.50 & 58.40 \\
$(1, 1)$ & 5.75 & 204.44 & 584.24 & 53.37 \\
$(1, 2)$ & 5.77 & 204.68 & 581.28 & 55.24 \\
\midrule
$(2, 1)$ -- MeEvo & \textbf{5.72} & \textbf{204.28} & \textbf{578.16} & \textbf{50.80} \\
\bottomrule
\end{tabular}%
}
\caption{Ablation results for different $(N, M)$ cycle configurations.}
\label{tab:ablation_nm}
\end{table}

\subsection{Computational Cost Analysis}
\label{subsec:cost}

\textbf{TABLE~\ref{tab:llm_cost}} reports approximate per-generation token consumption and wall-clock time for each method using DeepSeek-V4-Flash. Mean and standard deviation are computed over the two generation budgets ($L\in\{50,100\}$), and $\sim$ denotes values derived from approximate measurements. MeEvo uses approximately 65,000 tokens and 5.3 seconds per generation. Although this is the highest mean token cost, it is only 5\% above ReEvo (62,000) and 8\% above MCTS-AHD (60,000), while its standard deviation (9,000) is substantially lower than those of ReEvo (37,000), MCTS-AHD (23,000), and MeLA (23,000).

The corresponding DeepSeek-V4-Flash results in \textbf{Table~\ref{tab:comparative}} show that this overhead has different value across tasks. On the comparatively simple downstream-solver problems, TSP-ACO and BPP-ACO, MeEvo's advantage over MeLA is small (the latter has gaps of 1.96\% and 0.25\%, respectively). On the more constrained ACS and WSN problems, however, MeEvo obtains the lowest mean objectives and improves on MeLA by 1.79\% and 14.96\%, respectively. Thus, the additional per-generation cost is best interpreted as a task-dependent quality--cost trade-off: it is most justified when history-conditioned reflection produces meaningful gains on the harder constrained problems, rather than as a universal efficiency advantage.

\begin{table}[htbp]
\centering
\small
\caption{Per-generation LLM token cost and wall-clock time (\textit{DeepSeek-V4-Flash}). Mean and standard deviation are computed over $L\in\{50,100\}$; $\sim$ denotes values derived from approximate measurements.}
\label{tab:llm_cost}
\begin{tabular}{cccc}
\toprule
\textbf{Method} & \textbf{Mean} & \textbf{Std.} & \textbf{Time (s)} \\
\midrule
FunSearch & $\sim$15000 & $\sim$1240 & $\sim$1.5 \\
EoH & $\sim$16000 & $\sim$5800 & $\sim$1.6 \\
ReEvo & $\sim$62000 & $\sim$37000 & $\sim$2.3 \\
MCTS-AHD & $\sim$60000 & $\sim$23000 & $\sim$3.3 \\
MeLA & $\sim$45000 & $\sim$23000 & $\sim$4.2 \\
\textbf{MeEvo} & $\sim$65000 & $\sim$9000 & $\sim$5.3 \\
\bottomrule
\end{tabular}%
\vspace{2pt}
\end{table}

\subsection{Statistical Significance Analysis}
\label{subsec:significance}

We assess statistical significance using the one-sided Mann--Whitney $U$ test \cite{mann1947test}, comparing MeEvo against the second-best architecture within each LLM backbone on four problems ($n{=}5$ each; \textbf{TABLE~\ref{tab:mw_meevo}}). TSP-Construct is excluded because inter-architecture gaps are narrow (0.5--1.0 pp; \textbf{TABLE~\ref{tab:tsp_construct}}) and $n{=}5$ cannot separate architectures under these conditions.

\begin{table}[htbp]
\centering
\caption{Mann--Whitney $U$ Test: MeEvo vs.\ Second-Best Baseline (per LLM backbone)}
\vspace{2pt}
\small
\resizebox{\columnwidth}{!}{%
\begin{tabular}{cccccc}
\toprule
\multirow{2}{*}{Problem} & \multirow{2}{*}{2nd-Best Baseline} & \textbf{MeEvo} & \textbf{Baseline} & \multirow{2}{*}{$U$ Stat.} & \multirow{2}{*}{$p$-value$^{\dagger}$} \\
& & Mean($\downarrow$) $\pm$ Std($\downarrow$) & Mean($\downarrow$) $\pm$ Std($\downarrow$) & & \\
\midrule
\multicolumn{6}{c}{LLM backbone: \textit{DeepSeek-V4-Flash}} \\
\midrule
TSP-ACO & MCTS-AHD & $5.716 \pm 0.003$ & $5.812 \pm 0.016$ & $0$ & $0.004$ \\
BPP-ACO & MeLA & $204.28 \pm 0.10$ & $204.80 \pm 0.28$ & $1$ & $0.008$ \\
ACS & MeLA & $578.16 \pm 5.21$ & $588.50 \pm 9.07$ & $3$ & $0.028$ \\
WSN & MeLA & $50.80 \pm 0.76$ & $58.40 \pm 2.02$ & $0$ & $0.004$ \\
\midrule
\multicolumn{6}{c}{LLM backbone: \textit{MIMO-v2.5-Pro}} \\
\midrule
TSP-ACO & MCTS-AHD & $5.718 \pm 0.010$ & $5.807 \pm 0.010$ & $0$ & $0.004$ \\
BPP-ACO & MCTS-AHD & $204.40 \pm 0.24$ & $204.44 \pm 0.035$ & $10$ & $0.345$ \\
ACS & MeLA & $585.94 \pm 11.14$ & $609.47 \pm 15.21$ & $2$ & $0.016$ \\
WSN & MeLA & $52.00 \pm 2.26$ & $59.37 \pm 3.16$ & $0$ & $0.004$ \\
\bottomrule
\end{tabular}%
}
\vspace{2pt}
{\footnotesize $^{\dagger}$One-sided exact $p$-value. With $n_1{=}n_2{=}5$, the null distribution of $U$ is discrete on $\binom{10}{5}{=}252$ equally likely configurations; $U{=}0$ indicates complete separation (all 5 MeEvo runs below all 5 baseline runs).}
\label{tab:mw_meevo}
\end{table}

MeEvo achieves significant improvement in 7 of 8 comparisons at $\alpha{=}0.05$ (\textbf{TABLE~\ref{tab:mw_meevo}}), with complete separation ($U{=}0$, $p{=}0.004$) in four cases. The only non-significant result is BPP-ACO with MIMO ($p{=}0.345$), where the strong ACO structure leaves limited room for differentiation. These results confirm that MeEvo's advantage is systematic.

\section{Conclusion}

We introduced MeEvo, a framework that alternates Natural Evolution (crossover and mutation on heuristic code) with Metacognitive Evolution (reflection on the accumulated shared history). By treating reasoning traces and code as distinct entities with dedicated processing mechanisms, the cyclic architecture balances exploration and exploitation at two levels: process alternation governs the macro rhythm, while an autonomous shift from crossover to mutation handles the finer balance within Natural Evolution.

Across five problems, MeEvo consistently outperformed existing LLM-based AHD architectures under two independent LLM backbones. On the real-world constrained problems (ACS and WSN), where traditional metaheuristics largely failed, MeEvo succeeded with markedly lower variance, demonstrating robustness in domains where LLMs lack prior knowledge. On TSP-Construct, MeEvo remained competitive with other LLM-based methods across problem sizes and generalized reliably to larger instances, unlike deep RL approaches that collapse under distribution shift. Ablation confirmed that neither process alone suffices and that sufficient exploration before reflection yields the best performance. Significance testing validated that these gains are systematic.

Three limitations warrant discussion. First, MeEvo employs multiple LLM call types (crossover, mutation, metacognitive reflection, and error correction) whose individual contributions were not isolated; the overall computational cost scales with population size and evaluation budget. Second, the framework currently targets single-objective optimization; extending it to multi-objective settings remains an open challenge. Third, the operator balance uses a simple linear shift; whether a mechanism that adapts to convergence behavior could further improve performance warrants investigation.

Future work will address these limitations by exploring per-component contribution analysis, cost-efficient LLM calling strategies, multi-objective extensions, and convergence-aware operator balance mechanisms.

\bibliographystyle{IEEEtran}
\bibliography{reference}

\onecolumn
\appendix
\setcounter{table}{0}
\renewcommand{\thetable}{A\arabic{table}}

\subsection{Consolidated Notation and Symbol Definitions}
\label{app:notation}

\begin{table}[htbp]
\centering
\small
\renewcommand{\arraystretch}{1.35}
\caption{Consolidated Notation and Symbol Definitions}
\label{tab:notation}
\begin{tabular}{@{}ccc@{}}
\toprule
\textbf{Symbol} & & \textbf{Definition} \\
\midrule
\multicolumn{3}{c}{\textit{Problem and evaluation}} \\
\midrule
$P$ & & Target optimization problem \\
$APA$ & & Automated problem analysis description \\
$f(h), F$ & & Individual fitness and fitness set \\
$Err, ERR$ & & Individual and population error records \\
$Eval, L$ & & Evaluation counter and maximum budget \\
\midrule
\multicolumn{3}{c}{\textit{Population and evolution}} \\
\midrule
$N_0$ & & Initial population size \\
$N_{pop}$ & & Offspring count per generation \\
$h, H$ & & Individual heuristic and population set \\
$h^\ast$ & & Global best heuristic (elitist) \\
$\mathcal{P}$ & & Parent pool (top-$K$ feasible heuristics) \\
$K$ & & Parent pool size for truncation selection \\
$COR$ & & Error correction attempt limit \\
\midrule
\multicolumn{3}{c}{\textit{Reasoning-code representation}} \\
\midrule
$Th, TH$ & & Reasoning trace and reasoning history \\
$Meta$ & & Meta Insight (regulatory signal) \\
$\mathcal{R}$ & & Metacognitive reflection function (Eq.~\ref{eq:meta}) \\
$\mathcal{G}, \mathcal{T}$ & & Insight-to-code and insight-to-trace mapping functions (Eq.~\ref{eq:mapping}) \\
\midrule
\multicolumn{3}{c}{\textit{Cyclic alternation architecture}} \\
\midrule
$N, M$ & & Natural Evolution / Metacognitive Evolution cycle counts \\
$p_c, p_m$ & & Crossover and mutation probabilities \\
$\alpha$ & & Baseline crossover probability at convergence (Eq.~\ref{eq:pcross}) \\
$\beta$ & & Adaptation amplitude (Eq.~\ref{eq:pcross}) \\
\midrule
\multicolumn{3}{c}{\textit{Baseline method parameters (GP \& ACO)}} \\
\midrule
$\mu$ & & GP population size \\
$G$ & & GP generations \\
$r_c, r_m$ & & GP subtree crossover and mutation rates \\
$k_{tour}$ & & GP tournament selection size \\
$D_{\max}, d_{\max}$ & & GP max tree depth (evolution / initialization) \\
$e$ & & GP elitism count \\
$\gamma$ & & GP parsimony coefficient (penalizes tree size) \\
$n_{ants}, n_{iter}$ & & ACO colony size and iterations \\
$\rho$ & & ACO pheromone evaporation rate \\
$\alpha_{\!p}, \beta_{\!h}$ & & ACO pheromone / heuristic exponents (distinct from MeEvo $\alpha, \beta$) \\
$\beta^{x}, \beta^{y}$ & & WSN environmental fading coefficients (distinct from MeEvo $\beta$) \\
\bottomrule
\end{tabular}
\end{table}

\subsection{Auxiliary Algorithms}
\label{app:algorithms}

\setcounter{algorithm}{0}
\renewcommand{\thealgorithm}{A\arabic{algorithm}}

\begin{algorithm}[htbp]
\caption{Natural Evolution: Population-Driven Crossover and Mutation}
\label{alg:nature_evo}
\small
\begin{algorithmic}[1]
\setlength{\itemsep}{0pt}
\Procedure{NaturalEvo}{$H, F, Eval, L, N_{pop}, K$}
    \State \textbf{// Step 1: Select top-$K$ parents by fitness}
    \State $\mathcal{H}_{v} \gets \{h \in H \mid f(h) \neq \infty\}$
    \State $\mathcal{P} \gets$ top-$K$ of $\mathcal{H}_{v}$ sorted by $f(h)$ ascending
    \If{$|\mathcal{P}| < 2$}
        \State \Return $\emptyset, \emptyset$ \Comment{Insufficient parents}
    \EndIf
    \State \textbf{// Step 2: Compute dynamic probabilities}
    \State $p_c \gets \alpha + \beta \times (1 - Eval / L)$ \Comment{Crossover probability (Eq.~\ref{eq:pcross})}
    \State $p_m \gets 1 - p_c$ \Comment{Mutation probability}
    \State \textbf{// Step 3: Generate offspring via LLM}
    \State $H_{off} \gets \emptyset$, $TH_{off} \gets \emptyset$
    \For{$n \gets 1$ \textbf{to} $N_{pop}$}
        \State $r \gets \text{uniform}(0, 1)$
        \If{$r < p_c$}
            \State $h_a, h_b \gets$ random sample from $\mathcal{P}$, $2$ distinct
            \State $h_{off}, Th_{off} \gets \textsc{LLMCrossover}($
            \State \hspace{4em} $h_a, h_b, APA)$
        \Else
            \State $h_a \gets$ random sample from $\mathcal{P}$, $1$
            \State $h_{off}, Th_{off} \gets \textsc{LLMMutate}($
            \State \hspace{4em} $h_a, APA)$
        \EndIf
        \State $H_{off} \gets H_{off} \cup \{h_{off}\}$
        \State $TH_{off} \gets TH_{off} \cup \{Th_{off}\}$
    \EndFor
    \State \Return $H_{off}, TH_{off}$
\EndProcedure
\end{algorithmic}
\end{algorithm}

\begin{algorithm}[htbp]
\caption{Evaluation and Error Correction}
\label{alg:evaluate_repair}
\small
\begin{algorithmic}[1]
\setlength{\itemsep}{0pt}
\Procedure{EvalAndCorrect}{$h, Th, COR$}
    \State $ErrSet \gets \emptyset$
    \State $f(h), Err \gets \textsc{Evaluate}(h)$
    \If{$Err = \emptyset$}
        \State \Return $h, f(h), Th, ErrSet$
    \Else
        \State $ErrSet \gets ErrSet \cup \{Err\}$
        \State $h, f(h), Th_{R}, Err_{R} \gets$
        \State \hspace{1em} $\hyperref[alg:repair]{\textsc{ReplaceErr}}(h, Err, Th, COR)$
        \State $Th \gets Th \cup Th_{R}$
        \State $ErrSet \gets ErrSet \cup Err_{R}$
        \State \Return $h, f(h), Th, ErrSet$
    \EndIf
\EndProcedure
\end{algorithmic}
\end{algorithm}

\begin{algorithm}[htbp]
\caption{Replacement of Faulty Heuristics}
\label{alg:repair}
\small
\begin{algorithmic}[1]
\setlength{\itemsep}{0pt}
\Procedure{ReplaceErr}{$h_{Err}, Err, Th_{Err}, COR$}
    \State $h_{Rep} \gets h_{Err}$, $f(h_{Rep}) \gets \infty$
    \State $TH_{R} \gets \emptyset$, $ERR_{R} \gets \emptyset$
    \For{$k \gets 1$ \textbf{to} $COR$}
        \State $h_{C}, Th_{C} \gets \textsc{CorrectErrors}($
        \State \hspace{1em} $h_{Err}, Err, Th_{Err})$
        \State $TH_{R} \gets TH_{R} \cup \{Th_{C}\}$
        \State $f(h_{C}), Err_{C} \gets \textsc{Evaluate}(h_{C})$
        \If{$Err_{C} = \emptyset$ \textbf{and} $f(h_{C}) < f(h_{Rep})$}
            \State $h_{Rep} \gets h_{C}$, $f(h_{Rep}) \gets f(h_{C})$
        \ElsIf{$Err_{C} \neq \emptyset$}
            \State $ERR_{R} \gets ERR_{R} \cup \{Err_{C}\}$
        \EndIf
    \EndFor
    \State \Return $h_{Rep}, f(h_{Rep}), TH_{R}, ERR_{R}$
\EndProcedure
\end{algorithmic}
\end{algorithm}

\begin{algorithm}[htbp]
\caption{Metacognitive Reflection}
\label{alg:metaanalysis}
\small
\begin{algorithmic}[1]
\setlength{\itemsep}{0pt}
\Procedure{MetacognitiveReflection}{$TH, F, ERR, h^\ast$}
    \State $S_{f} \gets \textsc{AnalyzeFitness}(F, h^\ast)$
    \State $S_{o} \gets \textsc{RetainOptimal}(TH, F, h^\ast)$
    \State $S_{n} \gets \textsc{SearchBetter}(TH, F, ERR,$
    \State \hspace{4.5em} $S_{f}, S_{o})$
    \State $Meta \gets \textsc{ConstructPrompt}($
    \State \hspace{4.5em} $S_{f}, S_{o}, S_{n}, ERR, h^\ast)$
    \State \Return $Meta$
\EndProcedure
\end{algorithmic}
\end{algorithm}

\subsection{Prompt Details}
\begin{figure*}[htbp]
\begin{center}
    \begin{tcolorbox}[
        arc=3mm, 
        colback=white!5, 
        colframe=black, 
        boxrule=1pt, 
        width=\textwidth,
        before skip=0pt,       
        after skip=0pt,        
        boxsep=2pt,            
        left=5pt, right=5pt,   
        top=3pt, bottom=3pt,   
        enlarge top by=0pt,    
        enlarge bottom by=0pt, 
        before upper={\parindent=0pt\setlength{\parskip}{2pt}\linespread{0.8}\selectfont}
    ]
        ValueError: operands could not be broadcast together with shapes (50,) (50,150)\\
TypeError: 'numpy.float64' object is not callable\\
IndexError: too many indices for array: array is 1-dimensional, but 2 were indexed\\
ValueError: operands could not be broadcast together with shapes (50,) (50,150)\\
IndexError: invalid index to scalar variable.\\
ValueError: operands could not be broadcast together with shapes (50,150) (50,)\\
SyntaxError: '[' was never closed\\
SyntaxError: Generator expression must be parenthesized\\
SyntaxError: '[' was never closed\\
SyntaxError: '[' was never closed\\
ValueError: probabilities contain NaN\\
SyntaxError: closing parenthesis ']' does not match opening parenthesis '('\\
SyntaxError: '[' was never closed
    \end{tcolorbox}
    \caption{Examples of execution errors produced by LLM-generated heuristics under incomplete problem descriptions.}
    \label{Error}
\end{center}
\end{figure*}

\begin{figure*}[htbp]
\begin{center}
    \begin{tcolorbox}[
        arc=3mm, 
        colback=white!5, 
        colframe=black, 
        boxrule=1pt, 
        width=\textwidth,
        before skip=0pt,       
        after skip=0pt,        
        boxsep=2pt,            
        left=5pt, right=5pt,   
        top=3pt, bottom=3pt,   
        enlarge top by=0pt,    
        enlarge bottom by=0pt, 
        before upper={\parindent=0pt\setlength{\parskip}{2pt}\linespread{0.8}\selectfont}
    ]
        \textbf{Initial Description}\\
        Implement a function that performs adaptive course material recommendation using the SCSO metaheuristic optimization algorithm.

        \textcolor{green!70!black}{Heuristic operation success rate}
        : EOH - 5\% ReEvo - The operation rate is too low to proceed to the crossover and mutation stages.
        
        \textbf{Complete Description}\\         
        Implement a function that performs adaptive course material recommendation using the SCSO metaheuristic optimization algorithm. The function must:\\
        1. Strict Requirements for Parameter Access:\\
   - All parameters must be accessed using OBJECT.ATTRIBUTE notation only\\
   - NEVER use dictionary-style access like data\_al['lb'] or data\_al[0]\\
   - Required parameter accesses that must use dot notation:\\
   * data\_al.lb - lower bounds\\
     * data\_al.ub - upper bounds\\
     * data\_al.dim - problem dimension  \\
     * data\_al.SearchAgents - population size\\
     * data\_al.MaxIter - maximum iterations\\
     2. Mandatory Implementation Structure:\\
   \# 1. Parameter initialization (MUST use dot notation)\\
   lb = np.array(data\_al.lb)\\
   ub = np.array(data\_al.ub)\\
   dim = data\_al.dim\\
   SearchAgents\_no = data\_al.SearchAgents\\
   \# 2. Position initialization \\
   ub\_array = np.array(ub)\\
   lb\_array = np.array(lb)\\
   position initialization logic\\
   3. Core Algorithm Requirements:\\
   - Must implement both exploration and exploitation layers\\
   - Must include boundary constraint handling\\
   - Must use cosine-based position updates\\
   - Must maintain roulette wheel selection\\
4. Input/Output Specifications:\\
   Input Parameters:\\
   - data\_al: Algorithm config object with dot-accessible attributes\\
   - data\_pb: Problem data object\\
   - Positions: Current population positions\\
   - Best\_pos: Current best solution\\  
   - Best\_score: Current best fitness\\
   - rg: Current search radius\\
   Returns:\\
   - Updated Positions array only\\
   - NO other return values allowed\\
       \textcolor{green!70!black}{Heuristic operation success rate}
            : EOH - 75.49\% ReEvo - 40.71\%
    \end{tcolorbox}
    \caption{Incomplete problem descriptions will result in no runnable heuristics for both EoH and ReEvo.}
    \label{Descrip}
\end{center}
\end{figure*}

\begin{figure*}
\begin{center}
    \begin{tcolorbox}[
        arc=3mm,
        colback=white!5,
        colframe=black,
        boxrule=1pt,
        width=\textwidth,
        before skip=0pt,
        after skip=0pt,
        boxsep=4pt,
        left=8pt, right=8pt,
        top=5pt, bottom=5pt,
        before upper={\parindent=0pt\setlength{\parskip}{3pt}\linespread{0.95}\selectfont}
    ]
    \textbf{(1) Problem Analysis Expert} \hfill \textit{Function: \texttt{problem\_role()}}\\[2pt]
    Role: ``You are an expert in optimization algorithm design, metaheuristics, and computational complexity analysis.''\\[1pt]
    Task: Analyze the given optimization problem and produce a structured technical report.\\[1pt]
    Analysis framework:\\
    \quad (1) \textit{Problem Classification}---paradigm (combinatorial/continuous), search space (dimensionality, variable types, bounds), complexity class;\\
    \quad (2) \textit{Solution Approaches}---heuristic function role (inputs/computation/outputs), constraints (hard/soft with variable names), algorithm families (2--3 recommended with justification);\\
    \quad (3) \textit{Failure Mode Analysis}---2--3 specific failure modes with trigger conditions and vulnerable code regions.\\[1pt]
    Output: Numbered report (sections 1--3) with specific references to code elements.

    \vspace{4mm}\hrule\vspace{3mm}

    \textbf{(2) System Role---with problem context} \hfill \textit{Function: \texttt{system\_role(problem\_analysis)}}\\[2pt]
    Role: ``You are an expert in evolutionary computing, swarm intelligence, and algorithm optimization.''\\[1pt]
    Task: Using the embedded problem context, iteratively improve the provided algorithm. Reference the problem analysis to guide optimization choices.\\[1pt]
    Output format: A reasoning block \{thought\_process: (1) Error or weakness diagnosis, (2) Solution rationale, (3) Impact assessment\} followed by a complete Python code block with \#EVOLVE-START/\#EVOLVE-END markers.\\[1pt]
    Constraints: Modify ONLY code between \#EVOLVE-START/\#EVOLVE-END; keep function name \texttt{heuristics\_v2} and input signature unchanged; returned \texttt{Positions} must match baseline shape/type; no new dependencies beyond numpy.

    \vspace{4mm}\hrule\vspace{3mm}

    \textbf{(3) System Role---without problem context} \hfill \textit{Function: \texttt{system\_role()}}\\[2pt]
    Role: ``You are an expert in evolutionary computing and algorithm optimization.''\\[1pt]
    Task: Iteratively improve the provided algorithm by modifying only the designated region.\\[1pt]
    Output format and constraints: Identical to Variant (2), but without embedded problem context. Used when problem analysis is unavailable.

    \vspace{4mm}\hrule\vspace{3mm}

    \textbf{(4) Error Diagnosis Expert} \hfill \textit{Function: \texttt{error\_role()}}\\[2pt]
    Role: ``You are an expert Python code debugger specializing in numerical and scientific computing.''\\[1pt]
    Task: Fix the reported error in the provided algorithm code.\\[1pt]
    Output format: A reasoning block \{thought\_process: (1) Error diagnosis---classify root cause (syntax/type mismatch/shape error/index out of bounds/division by zero/logic flaw); (2) Solution rationale; (3) Impact assessment---confirm no breakage of other functionality\} followed by corrected Python code.\\[1pt]
    Constraints: Make the MINIMAL change necessary; do not refactor unrelated code; keep function name and signature unchanged.

    \vspace{4mm}\hrule\vspace{3mm}

    \textbf{(5) Metacognitive Reflection Expert} \hfill \textit{Function: \texttt{e\_learning\_role()}}\\[2pt]
    Role: ``You are a self-reflective AI optimizer conducting metacognitive analysis of your own evolutionary search process.''\\[1pt]
    Task: Examine the optimization history and identify strategic improvement directions.\\[1pt]
    Analysis framework:\\
    \quad (1) \textit{Cognitive Limitations}---systematic biases or blind spots (e.g., over-reliance on single search mechanism, insufficient diversity maintenance);\\
    \quad (2) \textit{Potential Biases}---optimization strategies repeated despite ineffectiveness, convergence to familiar patterns;\\
    \quad (3) \textit{Optimization Pathways}---specific algorithmic changes with named techniques (e.g., ``replace uniform random initialization with opposition-based learning'').\\[1pt]
    Constraints: Reference actual strategies from evolution history; keep analysis under 80 words; focus on structural changes, not parameter tuning.

    \end{tcolorbox}
    \caption{System-level role definitions for the five LLM call types in MeEvo. Each role is implemented as an XML-structured system prompt that defines the LLM's expertise persona, task specification, analysis framework, and output constraints. The corresponding user-level prompts are shown in \textbf{Fig.~\ref{pre}}.}
    \label{System}
\end{center}
\end{figure*}

\begin{figure*}
\begin{center}
    \begin{tcolorbox}[
        arc=3mm,
        colback=white!5,
        colframe=black,
        boxrule=1pt,
        fontupper=\footnotesize,
        width=\textwidth,
        before skip=0pt,
        after skip=0pt,
        boxsep=4pt,
        left=8pt, right=8pt,
        top=5pt, bottom=5pt,
        before upper={\parindent=0pt\setlength{\parskip}{3pt}\linespread{0.95}\selectfont}
    ]
    \textbf{(1) Problem Analysis Prompt} \hfill \textit{Function: \texttt{problem\_prompt(problem)}}\\[2pt]
    Task: Analyze the following optimization problem in depth.\\[1pt]
    Input: \{problem\}---the complete Python source code of the optimization problem.\\[1pt]
    Analysis requirements:\\
    \quad (1) \textit{Optimization Paradigm}---classify as combinatorial or continuous with evidence;\\
    \quad (2) \textit{Search Space}---dimensionality, variable types, bounds;\\
    \quad (3) \textit{Constraints}---enumerate all hard/soft constraints with variable names;\\
    \quad (4) \textit{Heuristic Function Role}---inputs, computation, outputs;\\
    \quad (5) \textit{Suitable Techniques}---3--5 proven techniques with justification;\\
    \quad (6) \textit{Failure Modes}---2--3 concrete failure modes with triggers.\\[1pt]
    Output: Numbered list (1--6), 2--4 sentences per section with specific code references.

    \vspace{4mm}\hrule\vspace{3mm}

    \textbf{(2) Initial Generation Prompt} \hfill \textit{Function: \texttt{return\_promoter\_init(...)}}\\[2pt]
    Role: Expert algorithm optimizer specializing in evolutionary computing and metaheuristics.\\[1pt]
    Inputs: \{problem\} (problem context), \{init\_code\} (baseline), \{init\_eval\} (history), \{code\} (current code).\\[1pt]
    Task: Transform the algorithm by selecting TWO complementary techniques from:\\
    \quad --- \textit{Exploration}: Levy flight, Cauchy mutation, opposition-based learning, chaotic initialization;\\
    \quad --- \textit{Exploitation}: local search (2-opt, VND), gradient-informed update, Nelder-Mead simplex;\\
    \quad --- \textit{Diversity}: crowding distance, fitness sharing, restart mechanism, adaptive population;\\
    \quad --- \textit{Constraint handling}: repair operator, feasibility rules, penalty adaptation.\\[1pt]
    Output: Reasoning block \{thought\_process\} + complete Python code. Modify ONLY between \#EVOLVE-START and \#EVOLVE-END.

    \vspace{4mm}\hrule\vspace{3mm}

    \textbf{(3) Error Correction Prompt} \hfill \textit{Function: \texttt{error\_prompt(str\_error, code\_str)}}\\[2pt]
    Task: Fix the reported error with minimal changes.\\[1pt]
    Inputs: \{str\_error\} (error message), \{code\_str\} (faulty code).\\[1pt]
    Requirements:\\
    \quad (1) \textit{Root Cause}---classify as syntax/type/shape/index/division/logic error with specific line;\\
    \quad (2) \textit{Fix}---minimal correction within \#EVOLVE-START/\#EVOLVE-END;\\
    \quad (3) \textit{Verification}---confirm no new issues introduced.\\[1pt]
    Output: \{error\_analysis\} + \{fix\_strategy\} + corrected Python code block.

    \vspace{4mm}\hrule\vspace{3mm}

    \textbf{(4) Metacognitive Reflection Prompt} \hfill \textit{Function: \texttt{e\_learning\_prompt(...)}}\\[2pt]
    Task: Perform metacognitive analysis of the evolutionary optimization process.\\[1pt]
    Inputs: \{thoughts\} (evolution reasoning records), \{errors\} (errors encountered), \{code\} (current best), \{history\} (convergence history).\\[1pt]
    Analysis framework:\\
    \quad (1) \textit{Convergence Diagnosis}---classify into PREMATURE\_CONVERGENCE / STEADY\_IMPROVEMENT / OSCILLATION / STAGNATION;\\
    \quad (2) \textit{Component Analysis}---identify 2--3 strongest algorithmic components with code references;\\
    \quad (3) \textit{Weakness Diagnosis}---1--2 fundamental weaknesses requiring structural redesign;\\
    \quad (4) \textit{Strategy Recommendation}---concrete change direction with named techniques.\\[1pt]
    Constraints: Do NOT repeat error-causing strategies; response under 100 words; reference actual code elements.

    \vspace{4mm}\hrule\vspace{3mm}

    \textbf{(5) Metacognition-Guided Generation Prompt} \hfill \textit{Function: \texttt{metacognition\_prompt(...)}}\\[2pt]
    Role: Expert algorithm optimizer performing an iteration of evolutionary improvement.\\[1pt]
    Inputs: \{problem\}, \{metacognition\} (metacognitive insights), \{init\_code\} (original baseline), \{init\_eval\}, \{code\} (current code).\\[1pt]
    Task: (1) Preserve strong components from the original baseline; (2) Redesign weak areas identified in metacognitive insights with fundamentally different approaches; (3) Modify ONLY between \#EVOLVE-START/\#EVOLVE-END; (4) Align with problem structure.\\[1pt]
    Output: \{reasoning\} (1--3 sentences on what insight guided the change) + complete Python code.

    \vspace{4mm}\hrule\vspace{3mm}

    \textbf{(6) Crossover Prompt} \hfill \textit{Function: \texttt{crossover\_prompt(...)}}\\[2pt]
    Role: Expert algorithm optimizer performing crossover between two parent solutions.\\[1pt]
    Inputs: \{parent1\_code\}, \{parent2\_code\}, \{problem\_description\}.\\[1pt]
    Task: (1) Analyze each parent's optimization technique and strengths; (2) Select superior component from each parent; (3) Synthesize a coherent hybrid resolving incompatibilities; (4) Introduce at least ONE novel enhancement.\\[1pt]
    Output: \{reasoning\} (2--4 sentences on component provenance and novel enhancement) + complete Python code.

    \vspace{4mm}\hrule\vspace{3mm}

    \textbf{(7) Mutation Prompt} \hfill \textit{Function: \texttt{mutation\_prompt(...)}}\\[2pt]
    Role: Expert algorithm optimizer performing an innovative mutation on a parent solution.\\[1pt]
    Inputs: \{parent\_code\}, \{problem\_description\}.\\[1pt]
    Task: (1) Analyze parent's primary optimization mechanism; (2) Identify 2--3 mutation opportunities; (3) Implement the most promising mutation that fundamentally changes exploration/exploitation.\\[1pt]
    Mutation guidance:\\
    \quad --- \textit{Continuous}: Levy flight, Cauchy mutation, adaptive step size, differential evolution, chaotic perturbation;\\
    \quad --- \textit{Combinatorial}: 2-opt/3-opt local search, swap/inversion mutation, simulated annealing acceptance, VND.\\[1pt]
    Output: \{reasoning\} (2--4 sentences on mutation type and expected impact) + complete Python code.

    \end{tcolorbox}
    \caption{Pre-defined user-level prompts for all seven LLM call types in MeEvo: (1) Problem Analysis, (2) Initial Generation, (3) Error Correction, (4) Metacognitive Reflection, (5) Metacognition-Guided Generation, (6) Crossover, (7) Mutation. Each prompt specifies the task, inputs, output format, and constraints. The corresponding system-level role definitions are shown in \textbf{Fig.~\ref{System}}.}
    \label{pre}
\end{center}
\end{figure*}

\clearpage

\subsection{Best Heuristic Codes Generated by MeEvo}
\label{app:best_code}

The following are the best heuristic codes generated by MeEvo (using DeepSeek-V4-Flash) for the four optimization problems. These codes correspond to the best objective values reported in \textbf{TABLE~\ref{tab:best_values}}.

\vspace{2mm}
\noindent\textbf{TSP-ACO} (Best Obj: 5.712)
\begin{lstlisting}[language=Python, basicstyle=\ttfamily\scriptsize, breaklines=true, breakatwhitespace=true, frame=single, numbers=left, numberstyle=\tiny\color{gray}, columns=fullflexible, keepspaces=true, showspaces=false]
def heuristics_v2(distance_matrix):
    dist = np.array(distance_matrix, dtype=float)
    n = dist.shape[0]
    eps = 1e-10
    inv_dist = 1.0 / (dist + eps)
    inv_max = np.max(inv_dist)
    inv_norm = inv_dist / inv_max if inv_max > 0 else inv_dist
    max_starts = min(250, max(40, n * 8))
    freq = np.zeros((n, n))
    all_costs = []
    tours = []
    for _ in range(max_starts):
        start = np.random.randint(n)
        unvisited = set(range(n))
        path = [start]
        unvisited.remove(start)
        current = start
        while unvisited:
            unvisited_list = np.array(list(unvisited))
            dists = dist[current, unvisited_list]
            probs = 1.0 / (dists + eps) ** 2.5
            probs /= probs.sum()
            next_city = np.random.choice(unvisited_list, p=probs)
            path.append(next_city)
            unvisited.remove(next_city)
            current = next_city
        tour = path
        tour_len = len(tour)
        improved = True
        max_iter = 200
        it = 0
        while improved and it < max_iter:
            improved = False
            it += 1
            for i in range(tour_len - 1):
                for j in range(i + 2, tour_len - (1 if i == 0 else 0)):
                    a, b = tour[i], tour[(i+1) % tour_len]
                    c, d = tour[j], tour[(j+1) % tour_len]
                    if dist[a, c] + dist[b, d] < dist[a, b] + dist[c, d] - 1e-9:
                        tour = tour[:i+1] + tour[i+1:j+1][::-1] + tour[j+1:]
                        improved = True
                        break
                if improved:
                    break
        cost = sum(dist[tour[i], tour[(i+1) % tour_len]] for i in range(tour_len))
        all_costs.append(cost)
        tours.append(tour)
    sorted_idx = np.argsort(all_costs)
    rank_weights = np.zeros(max_starts)
    for rank, idx in enumerate(sorted_idx):
        rank_weights[idx] = 1.0 / (rank + 1) ** 1.8
    for idx, tour in enumerate(tours):
        w = rank_weights[idx]
        for i in range(n):
            u = tour[i]
            v = tour[(i+1) % n]
            freq[u, v] += w
            freq[v, u] += w
    max_freq = np.max(freq)
    if max_freq > 0:
        norm_freq = freq / max_freq
    else:
        norm_freq = freq
    freq_pow = np.sqrt(norm_freq)
    alpha = 1.0
    heuristic = inv_norm * (1.0 + alpha * freq_pow)
    noise = 1.0 + 0.005 * np.random.randn(n, n)
    noise = (noise + noise.T) / 2.0
    heuristic = heuristic * noise
    K = max(7, min(45, n // 2))
    epsilon = 1e-6
    for i in range(n):
        row = heuristic[i, :]
        top = np.argsort(row)[-K:]
        mask = np.zeros(n, dtype=bool)
        mask[top] = True
        heuristic[i, ~mask] = epsilon
    heuristic = (heuristic + heuristic.T) / 2.0
    heuristic = np.clip(heuristic, epsilon, None)
    return heuristic
\end{lstlisting}

\vspace{2mm}
\noindent\textbf{BPP-ACO} (Best Obj: 204.20)
\begin{lstlisting}[language=Python, basicstyle=\ttfamily\scriptsize, breaklines=true, breakatwhitespace=true, frame=single, numbers=left, numberstyle=\tiny\color{gray}, columns=fullflexible, keepspaces=true, showspaces=false]
def heuristics_v2(node_attr, node_constraint):
    demand = np.asarray(node_attr, dtype=float)
    cap = np.asarray(node_constraint).flatten()[0]
    n = len(demand)
    i_idx = np.arange(n)[:, None]
    j_idx = np.arange(n)[None, :]
    sum_matrix = demand[i_idx] + demand[j_idx]
    feasible = (sum_matrix <= cap) & (i_idx != j_idx)
    n_feasible = feasible.sum(axis=1)
    prod_matrix = demand[i_idx] * demand[j_idx]
    waste = cap - sum_matrix
    heu = np.where(feasible, prod_matrix / (waste + 1e-10), 1e-10)
    crit_bonus = 1.0 / (n_feasible[i_idx] + n_feasible[j_idx] + 1.0)
    heu = heu * (1.0 + crit_bonus)
    np.fill_diagonal(heu, 0.0)
    return heu
\end{lstlisting}

\vspace{2mm}
\noindent\textbf{ACS} (Best Obj: 564.44)
\begin{lstlisting}[language=Python, basicstyle=\ttfamily\scriptsize, breaklines=true, breakatwhitespace=true, frame=single, numbers=left, numberstyle=\tiny\color{gray}, columns=fullflexible, keepspaces=true, showspaces=false]
def heuristics_v2(Positions, Best_pos, Best_score, rg):
    SearchAgents_no = Positions.shape[0]
    dim = Positions.shape[1]
    lb_array = np.zeros((SearchAgents_no, dim))
    ub_array = np.ones((SearchAgents_no, dim))
    rand_adjust = lb_array + (ub_array - lb_array) * np.random.rand(*Positions.shape)
    Positions = np.where((Positions < lb_array) | (Positions > ub_array), rand_adjust, Positions)
    if SearchAgents_no < 4:
        Positions = Positions + 0.1 * rg * np.random.randn(*Positions.shape) * (Best_pos - Positions)
        return np.clip(Positions, lb_array, ub_array)
    if np.any(np.isnan(Best_pos)) or np.isnan(Best_score):
        median_pos = np.median(Positions, axis=0)
        Positions = 0.5 * Positions + 0.5 * median_pos + 0.05 * np.random.randn(*Positions.shape)
        return np.clip(Positions, lb_array, ub_array)
    binary_Positions = (Positions > 0.5).astype(float)
    binary_Best = (Best_pos > 0.5).astype(float)
    archive_size = max(3, int(SearchAgents_no * (0.2 + 0.5 * (rg / 2.5))))
    def select_diverse_subset(pop, k):
        n = pop.shape[0]
        if n <= k:
            return np.arange(n)
        selected = [np.argmax(np.sum(np.abs(pop - binary_Best), axis=1))]
        for _ in range(k - 1):
            dists_to_selected = np.min([np.sum(np.abs(pop - pop[s]), axis=1) for s in selected], axis=0)
            dists_to_selected[selected] = -1
            selected.append(np.argmax(dists_to_selected))
        return np.array(selected)
    archive_indices = select_diverse_subset(binary_Positions, min(archive_size, SearchAgents_no))
    Archive = binary_Positions[archive_indices]
    idx = np.arange(SearchAgents_no)
    a1 = np.random.randint(0, len(Archive), size=SearchAgents_no)
    a2 = np.random.randint(0, len(Archive), size=SearchAgents_no)
    mask_same = (a1 == a2)
    a2 = np.where(mask_same, (a2 + 1) % len(Archive), a2)
    F = 0.5 + 0.4 * (rg / 2.5)
    binary_diff = Archive[a1] - Archive[a2]
    structural_diff = 4 * binary_diff * (1 - np.abs(binary_diff))
    mutant = Best_pos + F * structural_diff
    temperature = 0.05 + 0.95 * (rg / 2.5)
    uncertainty = np.abs(binary_Positions - Archive[a1]) * np.abs(binary_Positions - Archive[a2])
    uncertainty = np.clip(uncertainty + 0.05, 0, 1)
    CR_base = 0.5 + 0.4 * (1 - rg / 2.5)
    CR = CR_base + 0.35 * uncertainty * temperature
    CR = np.clip(CR, 0.2, 0.95)
    j_rand = np.random.randint(0, dim, size=SearchAgents_no)
    crossover_mask = np.random.rand(*Positions.shape) < CR
    crossover_mask[np.arange(SearchAgents_no), j_rand] = True
    trial = np.where(crossover_mask, mutant, Positions)
    proximity_to_threshold = 1 - np.abs(2 * trial - 1)
    push_strength = 0.5 * (1 - rg / 2.5) * proximity_to_threshold
    push_direction = (trial > 0.5).astype(float)
    trial = trial + push_strength * (push_direction - trial)
    current_binary = (trial > 0.5).astype(float)
    coord_importance = np.abs(trial - 0.5)
    direction_to_best = 2 * (binary_Best - 0.5)
    current_direction = 2 * (current_binary - 0.5)
    alignment = direction_to_best * (1 - 2 * (current_direction == direction_to_best).astype(float))
    explore_weight = rg / 2.5
    exploit_weight = 1 - explore_weight
    guided_score = coord_importance * (explore_weight * np.random.randn(*trial.shape) + exploit_weight * alignment)
    num_refine = max(1, int(dim * 0.05 * (2.5 - rg) / 2.5))
    if num_refine > 0:
        top_coords = np.argpartition(-np.abs(guided_score), num_refine, axis=1)[:, :num_refine]
        for i in range(SearchAgents_no):
            for j in top_coords[i]:
                if np.random.rand() < exploit_weight:
                    target = binary_Best[j]
                else:
                    target = 1 - current_binary[i, j]
                trial[i, j] = 0.5 + (target - 0.5) * (0.5 + 0.5 * np.random.rand())
    Positions = np.clip(trial, lb_array, ub_array)
    invalid_mask = ~np.isfinite(Positions)
    if np.any(invalid_mask):
        replacement = 0.5 + 0.2 * np.random.randn(*Positions.shape)
        Positions = np.where(invalid_mask, replacement, Positions)
        Positions = np.clip(Positions, lb_array, ub_array)
    return Positions
\end{lstlisting}

\vspace{2mm}
\noindent\textbf{WSN} (Best Obj: 50.00)
\begin{lstlisting}[language=Python, basicstyle=\ttfamily\scriptsize, breaklines=true, breakatwhitespace=true, frame=single, numbers=left, numberstyle=\tiny\color{gray}, columns=fullflexible, keepspaces=true, showspaces=false]
def heuristics_v2(Positions, Best_pos, Best_score, rg):
    SearchAgents_no = Positions.shape[0]
    dim = Positions.shape[1]
    num_cn = dim // 3
    lb = np.zeros(dim)
    ub = np.tile([50.0, 50.0, 30.0], 50)
    Positions = np.clip(Positions, lb, ub)
    F = 0.5 + 0.5 * (rg / 10.0)
    CR = 0.7 + 0.3 * (rg / 10.0)
    Cauchy_scale = 0.1 * (rg / 10.0)
    best_resh = Best_pos.reshape(num_cn, 3)
    def compute_connectivity_components(coords):
        dists = np.linalg.norm(coords[:, None] - coords[None, :], axis=2)
        adj = dists <= 20.0
        visited = np.zeros(num_cn, dtype=bool)
        components = []
        for i in range(num_cn):
            if not visited[i]:
                stack = [i]
                comp = []
                while stack:
                    node = stack.pop()
                    if visited[node]:
                        continue
                    visited[node] = True
                    comp.append(node)
                    neighbors = np.where(adj[node] & ~visited)[0]
                    stack.extend(neighbors.tolist())
                components.append(np.array(comp))
        return components
    def repair_connectivity(trial):
        trial_resh = trial.reshape(num_cn, 3)
        coords = trial_resh[:, :2]
        components = compute_connectivity_components(coords)
        if len(components) <= 1:
            return trial
        largest_idx = np.argmax([len(c) for c in components])
        largest_comp = components[largest_idx]
        other_nodes = np.concatenate([c for i, c in enumerate(components) if i != largest_idx])
        large_coords = coords[largest_comp]
        for node in other_nodes:
            node_coord = coords[node]
            dists = np.linalg.norm(large_coords - node_coord, axis=1)
            nearest_idx = largest_comp[np.argmin(dists)]
            trial_resh[node, :2] += 0.6 * (coords[nearest_idx] - node_coord)
            trial_resh[node, 2] = trial_resh[node, 2] * 0.4 + np.mean(trial_resh[largest_comp, 2]) * 0.6
        trial = np.clip(trial_resh.ravel(), lb, ub)
        return trial
    def repair_power(trial):
        trial_resh = trial.reshape(num_cn, 3)
        p = trial_resh[:, 2]
        std_p = np.std(p)
        if std_p > 1.0:
            mean_p = np.mean(p)
            p = mean_p + (p - mean_p) * (1.0 / std_p)
            trial_resh[:, 2] = p
            trial = np.clip(trial_resh.ravel(), lb, ub)
        return trial
    for i in range(SearchAgents_no):
        r1, r2 = np.random.randint(SearchAgents_no, size=2)
        while r1 == i:
            r1 = np.random.randint(SearchAgents_no)
        while r2 == i or r2 == r1:
            r2 = np.random.randint(SearchAgents_no)
        mutant = Positions[i] + F * (Best_pos - Positions[i]) + F * (Positions[r1] - Positions[r2])
        cauchy_noise = np.random.standard_cauchy(dim) * Cauchy_scale
        mutant += cauchy_noise
        j_rand = np.random.randint(dim)
        trial = np.where(np.random.rand(dim) < CR, mutant, Positions[i])
        trial[j_rand] = mutant[j_rand]
        trial = repair_connectivity(trial)
        trial = repair_power(trial)
        if rg < 2.0 and np.random.rand() < 0.3:
            pert = np.random.randn(dim) * (rg / 30.0)
            trial = Best_pos + pert
            trial = np.clip(trial, lb, ub)
            trial = repair_connectivity(trial)
            trial = repair_power(trial)
        Positions[i] = trial
    return Positions
\end{lstlisting}

\end{document}